\documentclass[11pt]{article}

\usepackage[preprint]{acl}

\usepackage{times}
\usepackage{latexsym}
\usepackage[T1]{fontenc}
\usepackage[utf8]{inputenc}
\usepackage{microtype}
\usepackage{inconsolata}
\usepackage{graphicx}
\usepackage{amsmath}
\usepackage{makecell}
\usepackage{amssymb}
\usepackage{booktabs}
\usepackage{multirow}
\usepackage{float}
\usepackage{algorithm}
\usepackage{algorithmic}
\usepackage{hyperref}

\title{ReNIO: Reweighting Negative Trajectory Importance for LLM On-Policy Distillation}

\author{
Chen Lin\textsuperscript{1} \quad
Kedi Chen\textsuperscript{1,2} \quad
Wei Zhang\textsuperscript{1,2}\Thanks{Corresponding author.} \\
\textsuperscript{1}East China Normal University \quad
\textsuperscript{2}Shanghai Innovation Institute \\
Shanghai, China \\
\texttt{\{clin,kdchen\}@stu.ecnu.edu.cn} \quad
\texttt{zhangwei.thu2011@gmail.com}
}

\usepackage{subcaption}
\usepackage{placeins}
\usepackage{xcolor}

\definecolor{dpos}{RGB}{25, 135, 84}
\definecolor{dneg}{RGB}{192, 57, 43}

\newcommand{\dsub}[2][pos]{%
  \llap{$\underset{%
    \textcolor{d#1}{\scalebox{0.6}{\textbf{#2}}}%
  }{\phantom{0}}$}%
}
\usepackage{colortbl}
\definecolor{lightblue}{rgb}{0.85,0.93,1.0}

\begin{document}
\maketitle

\begin{abstract}
On-policy distillation (OPD) improves LLM reasoning by training a student model on its own generated outputs, but standard OPD treats all student-generated outputs (SGOs) equally regardless of their informativeness. We observe a consistent asymmetry in controlled filtering experiments: in both OPD and on-policy self distillation (OPSD), training only on incorrect SGOs outperforms training only on correct ones. Our further analysis suggests that models trained on correct-only SGOs tend to generate shorter reasoning traces and show weaker reflection behavior, while incorrect SGOs better preserve exploratory reasoning near the model's capability boundary. To exploit this signal without requiring full answer-containing rollouts, we introduce ReNIO, which \textbf{Re}weights \textbf{N}egative trajectory \textbf{I}mportance for LLM \textbf{O}n-policy distillation. By using the student-to-teacher probability ratio, ReNIO identifies pivotal tokens leading to wrong reasoning traces  and aggregates their information into a normalized sample weight, inherently assigning larger weights to likely negative trajectories without observing the correctness of final-answer. Since ReNIO only uses prefix-conditioned token probabilities, it preserves OPD's prefix training advantage over full-rollout reinforcement learning. Across both mathematical reasoning and code generation tasks, ReNIO improves both OPD and OPSD, with representative relative gains of up to $8.90\%$ for Qwen3-1.7B and $10.00\%$ for R1-Distill-Qwen-7B on mathematical reasoning benchmarks. Code repo: \href{https://github.com/BDML-lab/ReNIO}{ https://github.com/BDML-lab/ReNIO}.
\end{abstract}

\section{Introduction}
Reinforcement learning (RL) and on-policy distillation (OPD) are becoming two mainstream paradigms for post-training large language models (LLMs) on reasoning tasks \citep{ke2026surveyfrontiersllmreasoning,chen2026surveyinductivereasoninglarge}. 
RL methods sample on-policy reasoning trajectories and update the model according to externally returned rewards~\cite{dapo, kimik2, deepseek-math, DSR1}. 
OPD, in contrast, trains on student-generated outputs (SGOs) by aligning the student's token-level distributions with teacher distributions along the same sampled trajectories~\cite{ThinkingMachineOPD, GKD}. 
This gives OPD a more dense supervision signal: while RL typically receives sparse sequence-level rewards and must address credit assignment over long trajectories, OPD can provide token-wise feedback at every visited prefix.

OPD also has a substantial training-cost advantage because it can learn from short SGO prefixes instead of full answer-containing rollouts with just little performance degradation ~\cite{zhang2026prefixopd, OPSD}. 
In comparison, reinforcement learning for reasoning tasks must observe the final answer of an on-policy reasoning trajectory before assigning a reward~\cite{RLVR}. 
For difficult reasoning problems, obtaining this answer may require generating very long trajectories, which makes RL substantially more expensive in generation, memory, and training time.

These two advantages make OPD attractive for efficient reasoning post-training, but they also shift the bottleneck from obtaining supervision to deciding which student-generated prefixes deserve more training emphasis. 
Standard OPD and on-policy self distillation(OPSD)\cite{OPSD} assign equal training weight to all SGOs, even though on-policy trajectories can differ greatly in how much useful learning signal they contain. 
Our controlled filtering experiments reveal a counterintuitive but consistent asymmetry: under both OPD and on-policy self-distillation (OPSD)~\cite{OPSD}, training only on incorrect SGOs consistently outperforms training only on correct SGOs. Further behavioral analysis shows that correct-only training tends to produce shorter responses with fewer reflection-style markers, whereas incorrect-only training better preserves exploratory and self-corrective reasoning. These findings suggest that negative trajectories can expose valuable process-level correction signals for the student's own failure modes, and may therefore play a more important role in OPD.

Hence, it's important to emphasize incorrect trajectories during OPD. The challenge is to emphasize informative negative trajectories without observing whether the final answer is correct. A direct correctness-based weighting rule would require rolling out each SGO until the answer appears, which would remove the short-prefix efficiency that makes OPD attractive. Moreover, the useful signal in an incorrect SGO is usually localized: a trajectory often becomes wrong because a few pivotal tokens steer the reasoning toward a wrong branch, while many surrounding tokens remain routine continuations. Therefore, an effective weighting signal should be prefix-computable.

We propose ReNIO, a sample-level reweighting method that emphasizes negative trajectories using only prefix-conditioned teacher and student probabilities. ReNIO identifies pivotal tokens by the student-to-teacher probability ratio: when the student assigns high probability to a token that the teacher assigns low probability, this ratio serves as prefix-computable evidence for a student-preferred but teacher-rejected branching decision. ReNIO selects such high-ratio tokens with a fixed threshold, aggregates their clipped log ratios by a geometric mean, and normalizes the resulting weights within each batch. The resulting objective redistributes training emphasis toward SGOs that contain strong corrective signals while preserving OPD's ability to train on truncated prefixes.

Our contributions are summarized as follows:
\begin{itemize}
    \item We identify a negative-trajectory asymmetry in OPD and OPSD: incorrect SGOs can be more useful than correct SGOs for improving reasoning performance.
    \item We introduce ReNIO, a prefix-computable sample reweighting method that uses student-to-teacher log ratios to emphasize likely negative, high-information trajectories without final-answer labels, thereby preserving the short-prefix training advantage of on-policy distillation without additional rollout or reward-labeling cost.
    \item We show that ReNIO improves both OPD and OPSD across mathematical reasoning and code generation tasks, with representative relative gains of up to $8.90\%$ for Qwen3-1.7B and $10.00\%$ for R1-Distill-Qwen-7B on mathematical reasoning benchmarks.
\end{itemize}

\section{Preliminary Analysis of On-Policy Distillation\label{section:motivtion experiments}}

We conduct controlled studies to understand which SGOs provide more useful OPD supervision. The results reveal two observations that motivate ReNIO's weighting strategy.
\label{sec:prelim}

\subsection{Incorrect-only Training Outperforms Correct-only Training}

A natural expectation is that correct SGOs should be the most useful training examples, since they contain successful reasoning paths from the student's own policy. Motivated by findings in reinforcement learning~\cite{negImportant}, we test this assumption on Qwen3-1.7B~\cite{Qwen3} by training OPD and OPSD variants using only correct or only incorrect SGOs. Additional setup details are provided in Appendix~\ref{sec:Detailed Experimental Setup}.

Figure~\ref{fig:a} ~\ref{fig:b} shows that incorrect-only training consistently outperforms correct-only training under both OPD and OPSD. Under OPD, the gains are $+3.60$, $+3.89$, and $+2.59$ Avg@12 points on AIME24, AIME25, and the benchmark average, respectively; under OPSD, the corresponding gains are $+1.94$, $+3.34$, and $+2.50$ points. Thus, correct and incorrect SGOs are not interchangeable, motivating a closer look at their inference-time effects.

\begin{figure*}[t]
\centering
\begin{subfigure}[t]{0.24\textwidth}
    \centering
    \includegraphics[width=\linewidth]{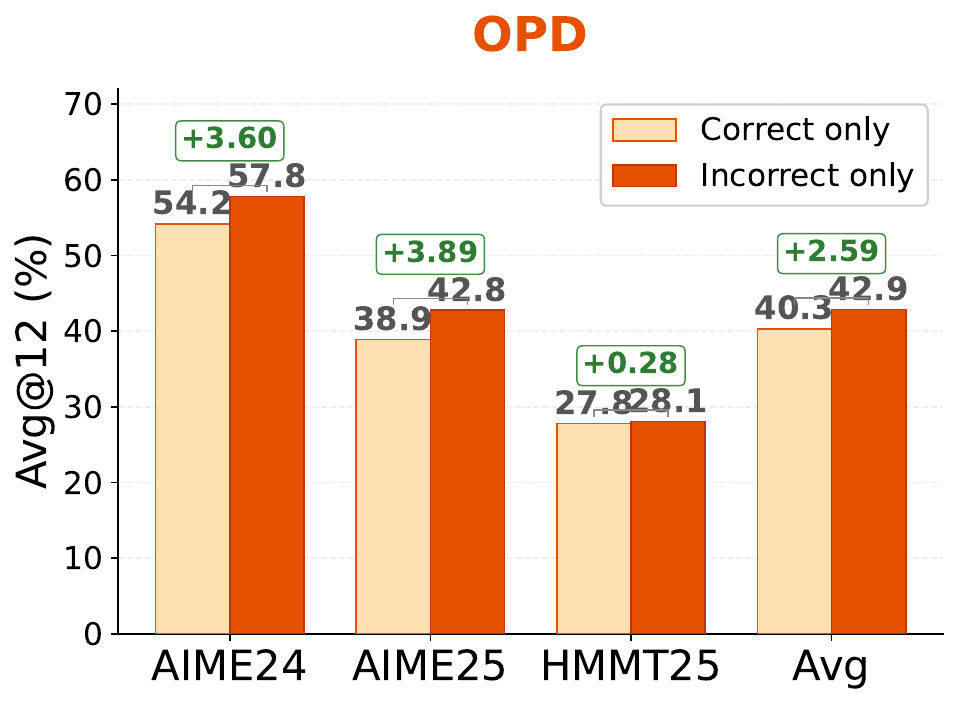}
    \caption{Correct-only vs. incorrect-only training under OPD.}
    \label{fig:a}
\end{subfigure}
\hfill
\begin{subfigure}[t]{0.24\textwidth}
    \centering
    \includegraphics[width=\linewidth]{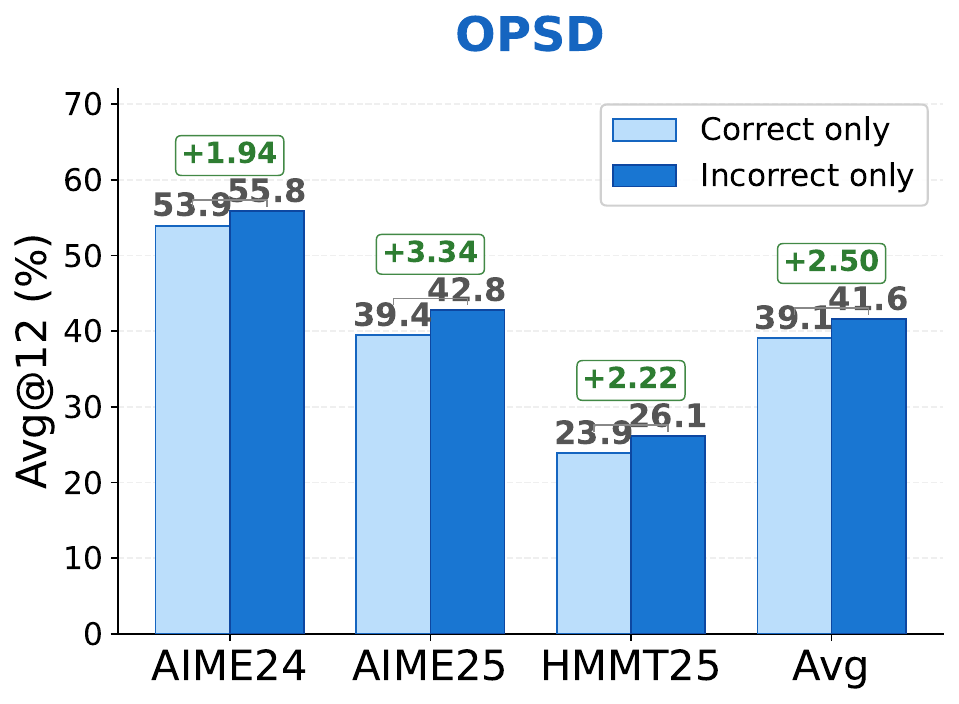}
    \caption{Correct-only vs. incorrect-only training under OPSD.}
    \label{fig:b}
\end{subfigure}
\hfill
\begin{subfigure}[t]{0.24\textwidth}
    \centering
    \includegraphics[width=\linewidth]{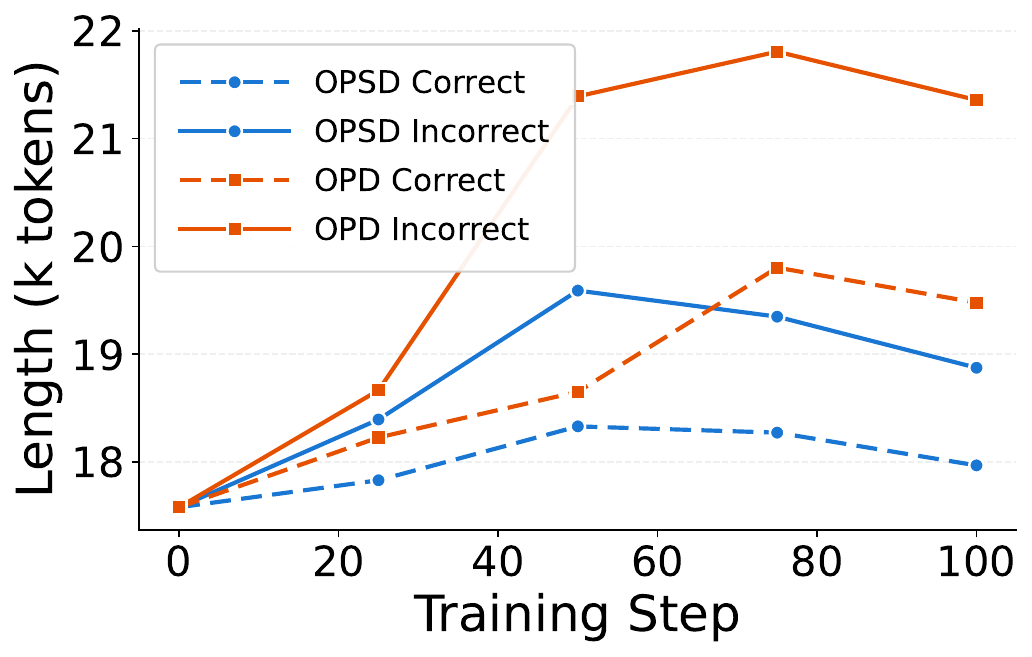}
    \caption{Average response length on AIME24.}
    \label{fig:c}
\end{subfigure}
\hfill
\begin{subfigure}[t]{0.24\textwidth}
    \centering
    \includegraphics[width=\linewidth]{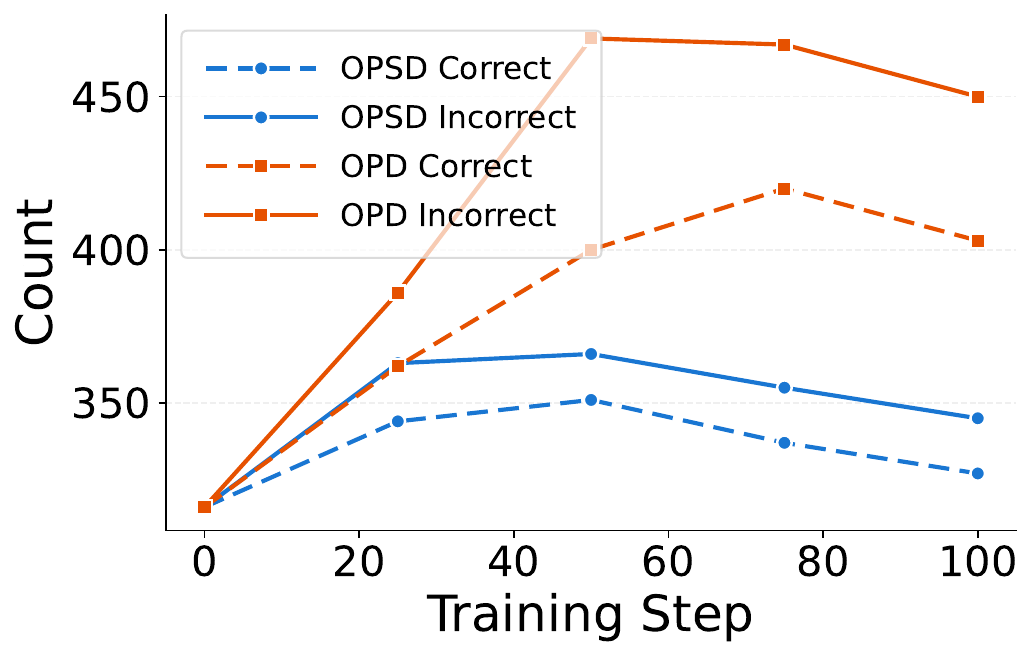}
    \caption{Average reflection-style marker count on AIME24.}
    \label{fig:d}
\end{subfigure}
\caption{Motivation analysis for correct-only versus incorrect-only on-policy distillation. Left two panels: Avg@12 (\%) under OPD and OPSD, where incorrect-only training consistently outperforms correct-only training across mathematical reasoning benchmarks. Right two panels: inference-time behavior on AIME24, where incorrect-only training produces longer responses and more reflection-style markers.}
\label{fig:correct_vs_incorrect}
\label{fig:gen_len}
\end{figure*}

\subsection{Length Patterns Suggest Different Training Roles\label{subsection:length experiments}}

To explain this gap, we evaluate the resulting models and record two inference-time statistics: average response length and the number of reflection-style markers. Following prior analyses of uncertainty and reasoning traces~\cite{whyOPSDNotWork, Kim2026UnderstandingRI}, we use the marker set \{``but'', ``wait'', ``hmm'', ``perhaps'', ``maybe'', ``actually'', ``seems'', ``might'', ``likely'', ``check'', ``alternatively'', ``however'', ``though'', ``again''\}.  Figure~\ref{fig:c} ~\ref{fig:d} shows AIME24 results; AIME25 and HMMT25 results are in Appendix~\ref{app:additional-experimental-results}.

Models trained on incorrect-only SGOs consistently generate longer responses than models trained on correct-only SGOs. Since length alone may reflect routine derivation rather than cautious reasoning, we also compare marker counts: incorrect-only training produces more reflection-style markers, indicating more frequent checking, revision, and alternative exploration.

These results suggest that \textcolor{blue}{correct-only training mainly consolidates behavior the student already performs successfully, but may also \textbf{make the model overly confident and less exploratory.}} \textcolor{red}{Incorrect-only training exposes failed but structured attempts from the current policy, helping \textbf{preserve cautious reasoning behavior}}. This interpretation is consistent with prior links between reasoning length, reflection markers, and robustness~\cite{whyOPSDNotWork}.

Standard OPD and OPSD nevertheless weight all trajectories equally. Our analysis suggests that trajectories exposing stronger student--teacher deviations deserve more emphasis, directly motivating ReNIO's sample-level weighting strategy.

\section{Methodology}

\begin{figure*}[t]
    \centering
    \begin{subfigure}[b]{\textwidth}
        \centering
        \includegraphics[width=\textwidth]{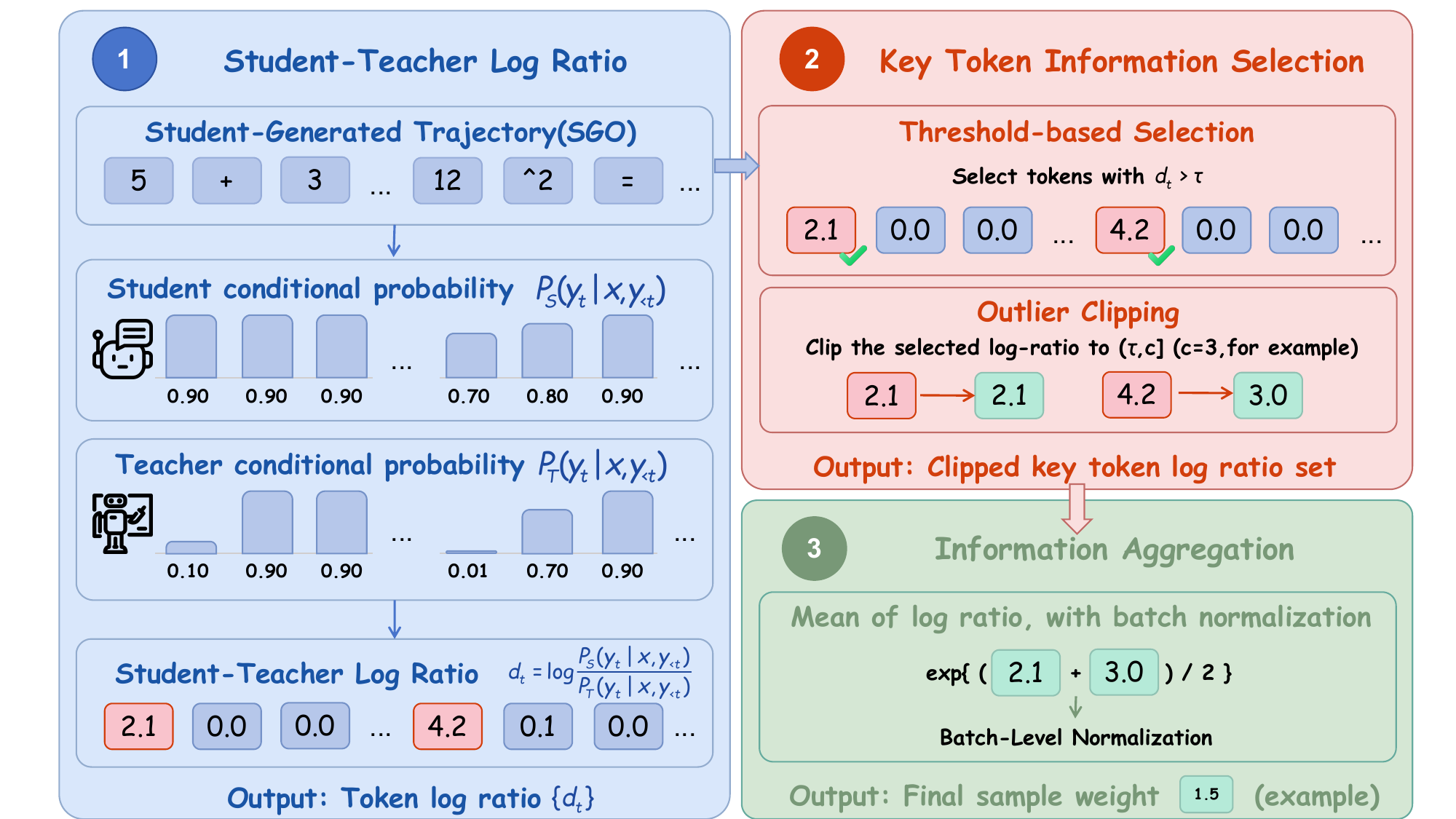}
        \caption{Overall pipeline of ReNIO.}
        \label{fig:pipeline_a}
    \end{subfigure}

    \begin{subfigure}[b]{\textwidth}
        \centering
        \includegraphics[width=\textwidth]{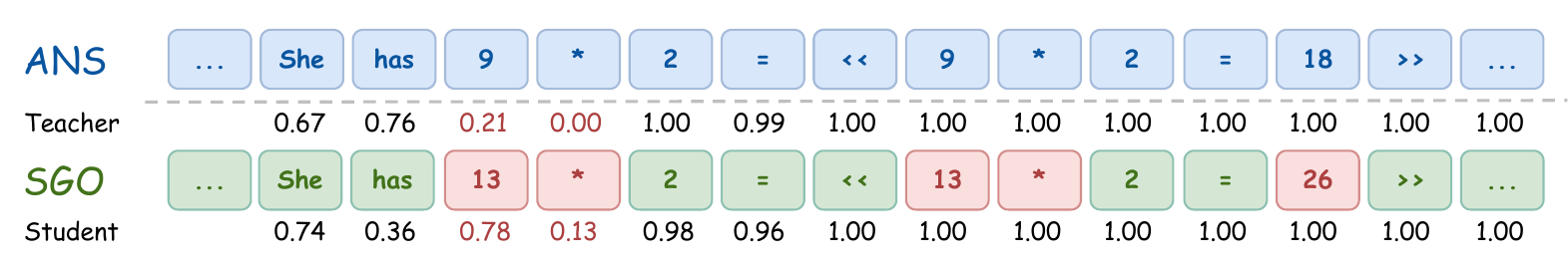}
        \caption{An example about how the pivotal tokens leading to a wrong trajectory, and teacher/student probilities on these tokens.}
        \label{fig:pipeline_b}
    \end{subfigure}
    
    \caption{Overview of the proposed ReNIO.\textcolor{blue}{Phase~I} computes the student--teacher log ratio along the SGO; \textcolor{red}{Phase~II} selects pivotal tokens with a fixed threshold; \textcolor{green!60!black}{Phase~III} aggregates selected log ratios into a normalized sample-level weight.}
    \label{fig:pipeline}
\end{figure*}

Figure~\ref{fig:pipeline} summarizes ReNIO. We first formalize the on-policy distillation setup (Section~\ref{sec:setup}), then describe three design components: a prefix-computable student--teacher log-ratio signal for SGO quality (Section~\ref{sec:pivotal}), fixed-threshold key-token selection (Section~\ref{sec:aggregate}), and geometric-mean aggregation into a normalized sample weight (Section~\ref{sec:weight}). Finally, we integrate the weight into the distillation objective (Section~\ref{sec:objective}).

\subsection{Problem Setup}
\label{sec:setup}

Consider a student policy $\pi_S$ and a teacher policy $\pi_T$. Given a prompt $x$ sampled from the training prompt distribution $D$, on-policy distillation first samples a student-generated output (SGO) $y=(y_1,\ldots,y_n)\sim \pi_S(\cdot|x)$, and then asks the teacher to provide token-level supervision on the same student-visited prefixes. For a model $M\in\{S,T\}$, we write
\begin{equation}
p_M^t(v)=\pi_M(v|x,y_{<t})
\end{equation}
for the next-token distribution at position $t$. We also use $\pi_M(y|x)$ as shorthand for the collection of prefix-conditioned distributions $\{p_M^t\}_{t=1}^{n}$ along the sampled SGO, rather than only the probability of the completed sequence.

Standard OPD trains the student by matching its prefix-conditioned distributions to the teacher's distributions on every sampled SGO. We define the sequence-level distillation divergence along $y$ as the sum of token-level divergences over all visited prefixes:
\begin{equation}
\mathcal{D}(\pi_S(y|x)\|\pi_T(y|x))
= \sum_{t=1}^{n} d(p_S^t,p_T^t),
\label{eq:distill}
\end{equation}
where $d(\cdot,\cdot)$ is the distribution divergence, like FKLD, RKLD. Thus, $\mathcal{D}$ does not denote a single divergence between two completed-sequence probabilities; it denotes the accumulated teacher--student discrepancy along the sampled SGO prefixes. The unweighted on-policy objective is therefore
\begin{equation}
L_{\mathrm{OPD}}
= \mathbb{E}_{x\sim D,\, y\sim \pi_S(\cdot|x)}
\bigl[\mathcal{D}(\pi_S(y|x)\|\pi_T(y|x))\bigr].
\label{eq:opd-objective}
\end{equation}
In this objective, every SGO contributes equally even though Section~\ref{sec:prelim} shows that different SGOs can affect learning in substantially different ways.

\subsection{Finding an Information Signal for SGO Quality}
\label{sec:pivotal}

The central challenge is to identify high-value SGOs without using their final answers. Section~\ref{sec:prelim} suggests that incorrect SGOs can play a distinct and useful role in OPD, but directly weighting trajectories by correctness would require complete answer-containing rollouts. ReNIO therefore looks for a prefix-computable signal that indicates when a student trajectory is likely to contain an informative failure.

The signal should focus on branching decisions rather than all tokens. A reasoning trajectory usually becomes wrong after a small number of local choices redirect an otherwise plausible solution path. In Figure~\ref{fig:pipeline_b}, for example, the student chooses ``13'' at a critical step. Because this token is sampled from the student policy, it has relatively high probability under $\pi_S(\cdot|x,y_{<t})$; because it moves the solution away from the teacher-preferred path, it has low probability under $\pi_T(\cdot|x,y_{<t})$. Such key tokens therefore have a large student-to-teacher probability ratio. After the wrong branch is taken, later tokens are conditioned on the same mistaken prefix, so teacher--student disagreement may shrink again. We therefore use the student-to-teacher ratio to locate the local decisions where the student is confident and the teacher rejects the sampled choice.

\paragraph{Token-level importance score.}
For each sampled token $y_t$ in an SGO, we define the student-to-teacher ratio
\begin{equation}
r_t \;=\; \frac{\pi_S(y_t\mid x,y_{<t})}{\pi_T(y_t\mid x,y_{<t})},
\label{eq:ratio}
\end{equation}
and its log form
\begin{equation}
\begin{aligned}
\ell_t
&= \log r_t \\
&= \log \pi_S(y_t|x,y_{<t})
 - \log \pi_T(y_t|x,y_{<t}).
\end{aligned}
\label{eq:logratio}
\end{equation}
A large $r_t$ means that the sampled token is much more likely under the student than under the teacher. We treat high-ratio tokens as \textbf{\emph{pivotal tokens}}: local decisions where the SGO departs from the teacher-preferred reasoning direction.

The log ratio also has an optimization interpretation. Under a prefix-level reverse-KL objective $\mathrm{KL}(p_S^t\|p_T^t)$, the effective gradient weight of a student-sampled token is $\log \pi_S(y_t|x,y_{<t})-\log \pi_T(y_t|x,y_{<t})=\ell_t$ up to a removable baseline (Appendix~\ref{app:rkl-gradient}). Thus, $\ell_t$ is both a disagreement signal and a token-level corrective weight. ReNIO uses this prefix-computable token signal as evidence for sample-level SGO importance, preserving OPD's ability to train on truncated prefixes without observing the final answer.

\subsection{Selecting Key Tokens from Long-Tailed Log Ratios}
\label{sec:aggregate}

The previous section shows that pivotal tokens should have large student-to-teacher ratios. Next step is to identify these tokens within a SGO. 

\begin{figure}[t]
\centering
\begin{subfigure}[b]{\columnwidth}
    \centering
    \includegraphics[width=\linewidth]{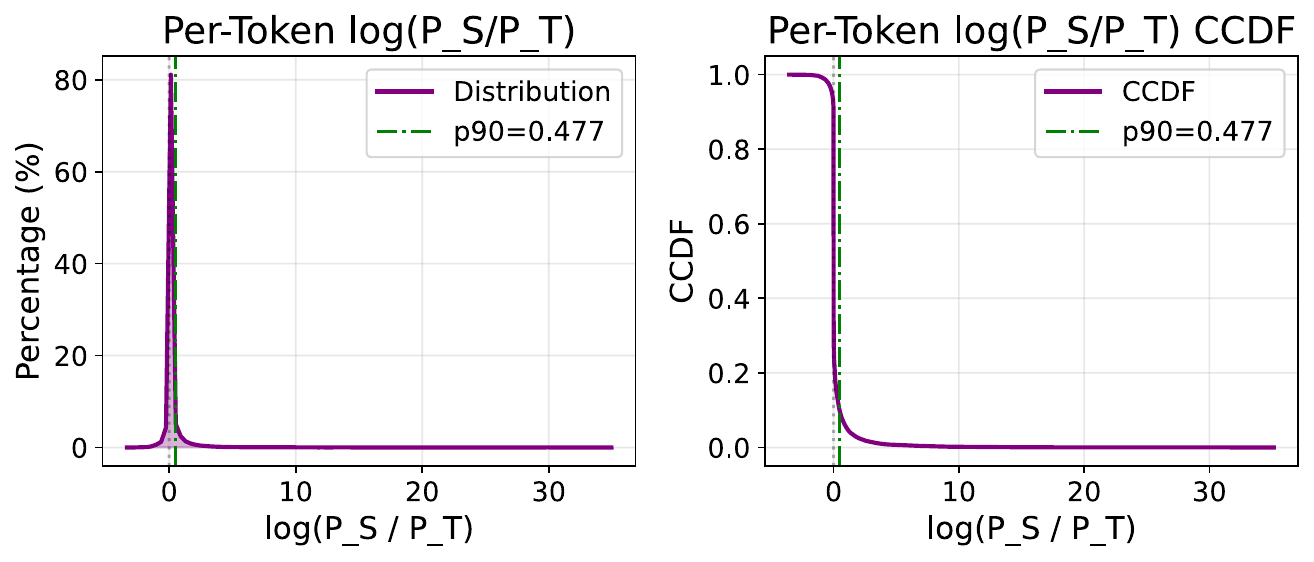}
    \caption{OPD}
    \label{fig:log_ratio_dist_opd}
\end{subfigure}
\\[2pt]
\begin{subfigure}[b]{\columnwidth}
    \centering
    \includegraphics[width=\linewidth]{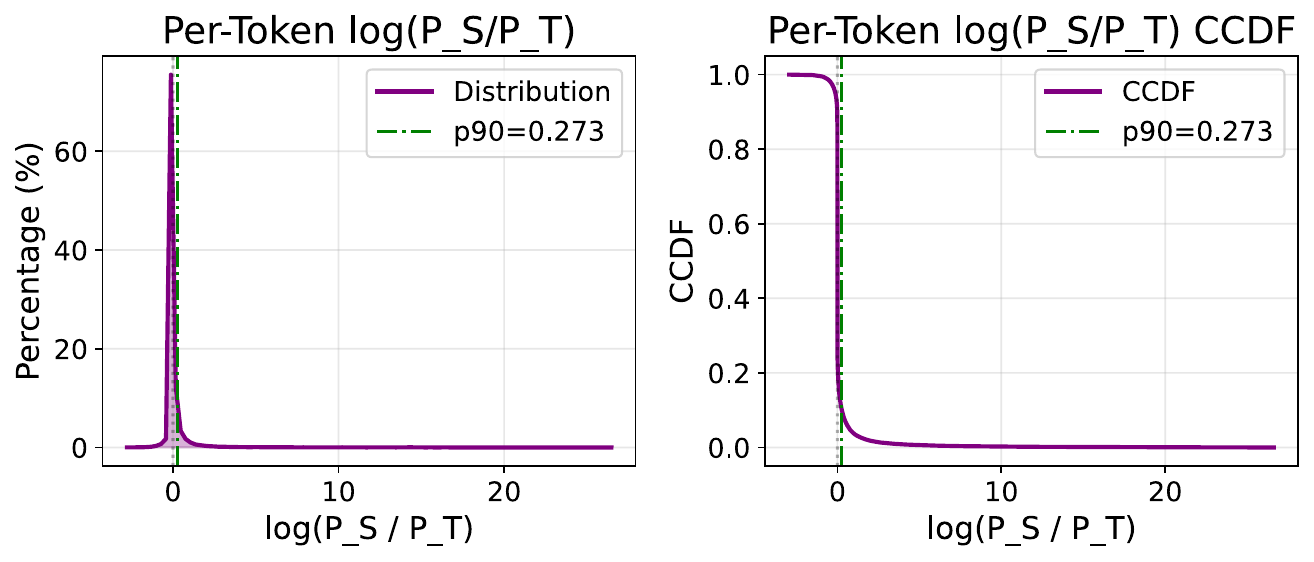}
    \caption{OPSD}
    \label{fig:log_ratio_dist_opsd}
\end{subfigure}
\caption{Distribution of token-level log ratios $\ell_t$ on SGOs in the Qwen3 mathematical reasoning setting. Both OPD and OPSD exhibit long-tailed distributions: most tokens have small log ratios, while a small number of tokens exhibit large student--teacher disagreement.}
\label{fig:log_ratio_dist}
\end{figure}

Figure~\ref{fig:log_ratio_dist} shows the empirical distribution of $\ell_t$ in the Qwen3 mathematical reasoning setting. Under both OPD and OPSD, the distribution is strongly long-tailed: most tokens have $\ell_t$ really close to zero, which means student and teacher align well on most tokens in SGO,  while only a small subset has much larger values. This pattern matches the intuition from Section~\ref{sec:pivotal}: \textcolor{red}{key tokens are really rare, but when they appear, the student-to-teacher ratio is large}.

This long-tailed structure motivates a simple fixed-threshold rule. ReNIO keeps a token as key token if its log ratio exceeds $\tau$, thus result in a key token set:
\begin{equation}
\mathcal{K}(x,y) = \bigl\{\, t \;\big|\; \ell_t > \tau \,\bigr\},
\label{eq:keyset}
\end{equation}
This rule removes the large mass of low-ratio tokens and retains the rare high-ratio tokens used to estimate SGO importance.

\subsection{From Key Tokens to Sample Weight}
\label{sec:weight}

After selecting $\mathcal{K}(x,y)$ for each SGO, we aggregate its token-level information into to get the final SGO weight. 
\paragraph{Geometric-mean aggregation.}
We use a \textbf{geometric mean}, equivalently averaging log student-teacher ratios and then exponentiating:
\begin{equation}
\begin{aligned}
w(x,y) 
&= \left( \prod_{t\in \mathcal{K}(x,y)} \frac{\pi_{S}(y_{t}|x_i,y_{<t})}{\pi_{T}(y_{t}|x_i,y_{<t})} \right)^{1/|\mathcal{K}(x,y)|}\\
&=\exp(\frac{1}{|\mathcal{K}(x,y)|}\sum_{t\in \mathcal{K}(x,y)}\log \frac{\pi_{S}(y_{t}|x_i,y_{<t})}{\pi_{T}(y_{t}|x_i,y_{<t})})\\
&=\exp(\frac{1}{|\mathcal{K}(x,y)|}\sum_{t\in \mathcal{K}(x,y)}\log r_t)
\end{aligned}
\label{eq:weight}
\end{equation}
when $\mathcal{K}(x,y)=\emptyset$, we will set $w(x,y)=1$.

\paragraph{Batch-level normalization.}
To keep the overall gradient scale unchanged, we normalize weights within each batch:
\begin{equation}
\hat{w}(x,y) = \frac{w(x,y)}{\bar{w}_B}, \quad \bar{w}_B = \frac{1}{B}\sum_{i=1}^{B} w(x_i, y_i),
\label{eq:norm}
\end{equation}
where $B$ is the batch size. The mean weight in each batch is therefore 1, so ReNIO redistributes emphasis across SGOs without changing the average update scale.


The complete weighting procedure is summarized in Algorithm~\ref{alg:weight} in Appendix~\ref{app:algorithm}.

\subsection{Weighted On-Policy Distillation Objective}
\label{sec:objective}

Finally we apply sample weights calculated by ReNIO to Equation ~\ref{eq:opd-objective}. The final objective is
\begin{equation}
\begin{aligned}
L_{\mathrm{ReNIO}} 
= & \mathbb{E}_{x\sim D, y\sim \pi_{S}(\cdot|x)}\\
&[\hat{w}(x,y)\mathcal{D}(\pi_{S}(y|x)||\pi_{T}(y|x))]\\
\end{aligned}
\label{eq:final}
\end{equation}
Since $\hat{w}(x,y)$ uses only prefix-conditioned token probabilities and does not require the final answer, ReNIO remains compatible with truncated-prefix training and preserves the efficiency advantage of OPD over reward-based RL methods.

\begin{table*}[t]
\centering
\small
\setlength{\tabcolsep}{5.5pt}
\renewcommand{\arraystretch}{1.08}
\begin{tabular}{lccccccc}
\toprule
\multirow{2}{*}{Model}
  & \multicolumn{4}{c}{Math Tasks}
  & \multicolumn{3}{c}{Coding Tasks} \\
\cmidrule(lr){2-5}\cmidrule(lr){6-8}
  & AIME24 & AIME25 & HMMT25 & Avg
  & HumanEval+ & MBPP+ & Avg \\
\midrule
\rowcolor{black!4}
  Qwen3-1.7B
    & 51.33 & 36.00 & 23.33 & 36.89
    & 73.63 & 57.21 & 65.42 \\
  w/ GRPO
    & 49.44 & 38.61 & 23.61 & 37.22
    & 74.84 & \textbf{65.87} & 70.36 \\
  w/ OPSD
    & 53.06 & 41.94 & 27.50 & 40.83
    & 74.85 & 62.43 & 68.64 \\
  \cellcolor{lightblue}w/ OPSD+ReNIO
    & \cellcolor{lightblue}\textbf{57.78}\dsub{+8.90\%} & \cellcolor{lightblue}\textbf{42.78}\dsub{+2.00\%} & \cellcolor{lightblue}\textbf{27.78}\dsub{+1.02\%} & \cellcolor{lightblue}\textbf{42.78}\dsub{+4.77\%}
    & \cellcolor{lightblue}\textbf{77.90}\dsub{+4.07\%} & \cellcolor{lightblue}63.16\dsub{+1.17\%} & \cellcolor{lightblue}\textbf{70.53}\dsub{+2.75\%} \\
\midrule
\rowcolor{black!4}
  Qwen3-4B
    & 74.44 & 65.28 & 42.22 & 60.65
    & 86.74 & 77.31 & 82.03 \\
  w/ GRPO
    & 73.89 & 67.78 & 42.22 & 61.30
    & 86.28 & 77.78 & 82.03 \\
  w/ OPSD
    & 75.56 & 68.89 & 43.33 & 62.59
    & 86.43 & 77.78 & 82.11 \\
  \cellcolor{lightblue}w/ OPSD+ReNIO
    & \cellcolor{lightblue}\textbf{75.83}\dsub{+0.36\%} & \cellcolor{lightblue}\textbf{69.17}\dsub{+0.41\%} & \cellcolor{lightblue}\textbf{46.67}\dsub{+7.71\%} & \cellcolor{lightblue}\textbf{63.89}\dsub{+2.08\%}
    & \cellcolor{lightblue}\textbf{87.04}\dsub{+0.71\%} & \cellcolor{lightblue}\textbf{78.31}\dsub{+0.68\%} & \cellcolor{lightblue}\textbf{82.68}\dsub{+0.69\%} \\
\midrule
\rowcolor{black!4}
  Qwen3-8B
    & 74.17 & \textbf{71.67} & 44.72 & 63.52
    & 85.52 & 77.84 & 81.68 \\
  w/ GRPO
    & \textbf{78.33} & 69.44 & 44.72 & 64.16
    & 87.20 & \textbf{78.51} & \textbf{82.86} \\
  w/ OPSD
    & 75.28 & 71.39 & 45.56 & 64.08
    & 86.59 & 77.05 & 81.82 \\
  \cellcolor{lightblue}w/ OPSD+ReNIO
    & \cellcolor{lightblue}78.06\dsub{+3.69\%} & \cellcolor{lightblue}71.11\dsub[neg]{-0.39\%} & \cellcolor{lightblue}\textbf{47.50}\dsub{+4.26\%} & \cellcolor{lightblue}\textbf{65.56}\dsub{+2.31\%}
    & \cellcolor{lightblue}\textbf{88.41}\dsub{+2.10\%} & \cellcolor{lightblue}77.31\dsub{+0.34\%} & \cellcolor{lightblue}\textbf{82.86}\dsub{+1.27\%} \\
\midrule
\rowcolor{black!4}
  R1-Distill-Qwen-1.5B
    & 27.78 & 22.78 & 14.44 & 21.67
    & 49.09 & 41.07 & 45.08 \\
  w/ GRPO
    & \textbf{31.67} & 24.17 & 13.89 & \textbf{23.24}
    & 49.24 & 41.47 & 45.36 \\
  w/ OPSD
    & 26.66 & 23.89 & 13.06 & 21.20
    & \textbf{50.46} & 40.87 & 45.67 \\
  \cellcolor{lightblue}w/ OPSD+ReNIO
    & \cellcolor{lightblue}28.61\dsub{+7.31\%} & \cellcolor{lightblue}\textbf{24.44}\dsub{+2.30\%} & \cellcolor{lightblue}\textbf{15.28}\dsub{+17.00\%} & \cellcolor{lightblue}22.78\dsub{+7.45\%}
    & \cellcolor{lightblue}50.30\dsub[neg]{-0.32\%} & \cellcolor{lightblue}\textbf{42.13}\dsub{+3.08\%} & \cellcolor{lightblue}\textbf{46.22}\dsub{+1.20\%} \\
\midrule
\rowcolor{black!4}
  R1-Distill-Qwen-7B
    & \textbf{56.11} & 40.83 & 25.56 & 40.83
    & 73.17 & 58.73 & 65.95 \\
  w/ GRPO
    & 54.17 & 40.83 & 24.17 & 39.72
    & 74.70 & 58.27 & 66.49 \\
  w/ OPSD
    & 55.56 & 38.89 & 24.61 & 39.69
    & 73.93 & 59.26 & 66.60 \\
  \cellcolor{lightblue}w/ OPSD+ReNIO
    & \cellcolor{lightblue}55.83\dsub{+0.49\%} & \cellcolor{lightblue}\textbf{42.78}\dsub{+10.00\%} & \cellcolor{lightblue}\textbf{26.11}\dsub{+6.09\%} & \cellcolor{lightblue}\textbf{41.57}\dsub{+4.74\%}
    & \cellcolor{lightblue}\textbf{74.85}\dsub{+1.24\%} & \cellcolor{lightblue}\textbf{59.39}\dsub{+0.22\%} & \cellcolor{lightblue}\textbf{67.12}\dsub{+0.78\%} \\
\bottomrule
\end{tabular}
\caption{Results of methods that do not require an additional teacher model.
  \textbf{Math Tasks}: AIME24, AIME25, and HMMT25 (Avg@12).
  \textbf{Coding Tasks}: HumanEval+ and MBPP+ (Avg@4).
  \textbf{Bold} denotes the best result per column within each model block.
  Subscripts on ReNIO rows report relative changes over the corresponding non-ReNIO row.}
\label{tab:main-no-teacher}
\end{table*}

\begin{table*}[t]
\centering
\small
\setlength{\tabcolsep}{5.5pt}
\renewcommand{\arraystretch}{1.08}
\begin{tabular}{lccccccc}
\toprule
\multirow{2}{*}{Model}
  & \multicolumn{4}{c}{Math Tasks}
  & \multicolumn{3}{c}{Coding Tasks} \\
\cmidrule(lr){2-5}\cmidrule(lr){6-8}
  & AIME24 & AIME25 & HMMT25 & Avg
  & HumanEval+ & MBPP+ & Avg \\
\midrule
\multicolumn{8}{c}{\emph{Teacher: Qwen3-8B}} \\
\rowcolor{black!4}
  Qwen3-1.7B
    & 51.33 & 36.00 & 23.33 & 36.89
    & 73.63 & 57.21 & 65.42 \\
  w/ OPD
    & 54.44 & 39.16 & 27.50 & 40.37
    & 74.40 & 62.43 & 68.42 \\
  \cellcolor{lightblue}w/ OPD+ReNIO
    & \cellcolor{lightblue}\textbf{54.72}\dsub{+0.51\%} & \cellcolor{lightblue}\textbf{43.06}\dsub{+9.96\%} & \cellcolor{lightblue}\textbf{28.33}\dsub{+3.02\%} & \cellcolor{lightblue}\textbf{42.04}\dsub{+4.14\%}
    & \cellcolor{lightblue}\textbf{75.00}\dsub{+0.81\%} & \cellcolor{lightblue}\textbf{63.56}\dsub{+1.81\%} & \cellcolor{lightblue}\textbf{69.28}\dsub{+1.26\%} \\
\midrule
\multicolumn{8}{c}{\emph{Teacher: DS-R1-Qwen-7B}} \\
\rowcolor{black!4}
  DS-R1-Qwen-1.5B
    & 27.78 & 22.78 & 14.44 & 21.67
    & 49.09 & 41.07 & 45.08 \\
  w/ OPD
    & 28.89 & 23.06 & 14.44 & 22.13
    & 50.15 & 41.14 & 45.65 \\
  \cellcolor{lightblue}w/ OPD+ReNIO
    & \cellcolor{lightblue}\textbf{29.72}\dsub{+2.87\%} & \cellcolor{lightblue}\textbf{23.89}\dsub{+3.60\%} & \cellcolor{lightblue}\textbf{16.67}\dsub{+15.44\%} & \cellcolor{lightblue}\textbf{23.43}\dsub{+5.87\%}
    & \cellcolor{lightblue}\textbf{50.91}\dsub{+1.52\%} & \cellcolor{lightblue}\textbf{41.20}\dsub{+0.15\%} & \cellcolor{lightblue}\textbf{46.06}\dsub{+0.90\%} \\
\bottomrule
\end{tabular}
\caption{Results of methods that require an additional teacher model.}
\label{tab:main-teacher}
\end{table*}

\section{Experiments\label{section: Main Experiments}}

\subsection{Experimental Setup}

\paragraph{Models and distillation settings.}
We evaluate ReNIO on two model families: Qwen3~\cite{Qwen3} and DeepSeek-R1-Distill-Qwen~\cite{ds_distill_qwen}. We study both teacher-based OPD and teacher-free OPSD. For OPD, we use Qwen3-1.7B and DeepSeek-R1-Distill-Qwen-1.5B as students, with Qwen3-8B and DeepSeek-R1-Distill-Qwen-7B as the corresponding teachers. For OPSD, we evaluate Qwen3-1.7B, Qwen3-4B, Qwen3-8B, DeepSeek-R1-Distill-Qwen-1.5B, and DeepSeek-R1-Distill-Qwen-7B. 

\paragraph{Tasks and training data.}
We evaluate ReNIO on two task domains: mathematical reasoning and code generation. For mathematical reasoning, we follow the data construction protocol of OPSD~\cite{OPSD} and use the same math training data for fair comparison. For code generation, we sample 30k coding-related examples from OpenThoughts~\cite{Openthoughts}. 

\paragraph{Evaluation.}
We evaluate mathematical reasoning on AIME2024~\cite{aime24}, AIME2025~\cite{aime25}, and HMMT2025~\cite{hmmt25}, reporting the average pass@$12$ (Avg@12) across the three benchmarks. Using evalplus\cite{evalplus}, we evaluate code generation on HumanEval+~\cite{evalplus} and MBPP+~\cite{evalplus}, reporting average pass@$4$ (Avg@4).For OPSD and OPD, we evaluate checkpoints every 25 steps up to 100 steps and report the best score. For GRPO, we report the peak performance within 300 training steps.

\paragraph{Implementation details.}
Following previous work\cite{OPSD}, for OPSD, we fix the teacher parameters to the initial model parameters and use full-vocabulary logit distillation and LoRA for all experiments. We set the maximum generation length to 1024 for Qwen3 OPD and OPSD training and 2048 for DeepSeek-R1-Distill-Qwen OPD and OPSD training; thus, training uses only SGO prefixes.
Detailed information is provided in Appendix~\ref{sec:Detailed Experimental Setup}.

\subsection{Main Results}
\label{subsec:main-results}


Tables~\ref{tab:main-no-teacher} and~\ref{tab:main-teacher} report the main results on mathematical reasoning and code generation. ReNIO consistently improves the corresponding OPD or OPSD baseline across all model blocks in terms of average performance: for OPSD, DS-R1-Qwen-7B obtains a 10.00\% relative improvement on AIME25; for OPD, DS-R1-Qwen-1.5B obtains a 15.44\% relative improvement on HMMT25, showing that reweighting negative on-policy trajectories is beneficial for both teacher-free self-distillation and teacher-based distillation.

Meanwhile, for DS-R1-Qwen-1.5B, standard OPSD slightly decreases the math average compared with the base model, while OPSD+ReNIO improves over both OPSD and the base model on all three math benchmarks. This suggests that ReNIO not only improves average performance, but also stabilizes on-policy self-distillation when the unweighted OPSD signal is noisy.

\begin{figure}[t]
\centering
\includegraphics[width=0.9\columnwidth]{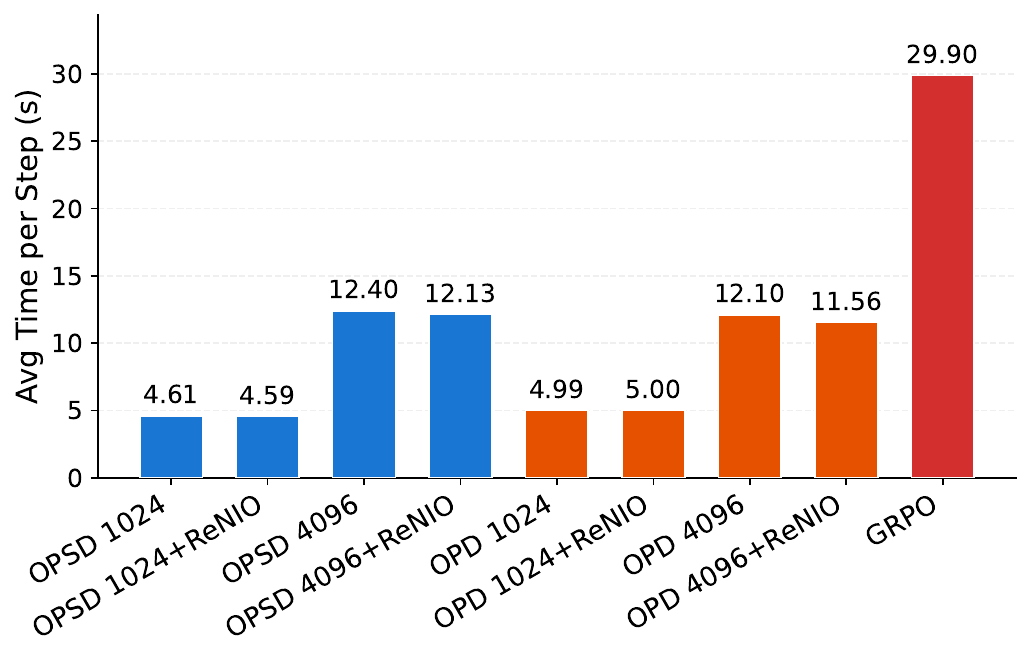}
\caption{Training-time comparison between GRPO and prefix-based on-policy distillation on Qwen3-1.7B mathematical reasoning.}
\label{fig:training_time}
\end{figure}

\subsection{Effective and Efficient Training from Short SGO Prefixes}
\label{subsec:efficiency-effectiveness}

The ability to train on short SGO prefixes is a central practical advantage of OPD and OPSD. Unlike GRPO, which must generate a complete trajectory before assigning a reward, prefix-based distillation can apply dense teacher supervision before the final answer appears. Figure~\ref{fig:training_time} shows the resulting cost difference: OPD and OPSD with 1024-token prefixes are substantially cheaper than GRPO, while extending the prefix length from 1024 to 4096, which usually covers the final answer in this setting, nearly triples the training time.

Table~\ref{tab:prefix-length-opsd} shows that this efficiency does not require a large loss in effectiveness. Without ReNIO, 1024-token OPSD even outperforms 4096-token OPSD on average (40.83 vs. 38.70), and increasing OPD from 1024 to 4096 tokens improves the average only modestly (40.37 to 41.38). ReNIO further improves average performance under both prefix lengths: OPSD gains 1.95 points at 1024 tokens and 1.86 points at 4096 tokens, while OPD gains 1.67 and 0.93 points, respectively. Because ReNIO computes weights from prefix-conditioned student and teacher probabilities, these gains do not require answer-containing rollouts. These results therefore support two conclusions: \textbf{short prefixes already provide effective OPD supervision, and ReNIO preserves this low-cost regime while improving it.}

\begin{table}[t]
\centering
\small
\setlength{\tabcolsep}{4.5pt}
\renewcommand{\arraystretch}{1.08}
\begin{tabular}{lcccc}
\toprule
Method & AIME24 & AIME25 & HMMT25 & Avg \\
\midrule
OPSD(1024) & 53.06 & 41.94 & 27.50 & 40.83 \\
\cellcolor{lightblue}/w ReNIO
    & \cellcolor{lightblue}\textbf{57.78}
    & \cellcolor{lightblue}42.78
    & \cellcolor{lightblue}\textbf{27.78}
    & \cellcolor{lightblue}\textbf{42.78} \\

OPSD(4096) & 53.33 & 37.22 & 25.56 & 38.70 \\
\cellcolor{lightblue}/w ReNIO
    & \cellcolor{lightblue}53.33
    & \cellcolor{lightblue}\textbf{43.06}
    & \cellcolor{lightblue}25.28
    & \cellcolor{lightblue}40.56 \\

\midrule
OPD(1024) & 54.44 & 39.16 & 27.50 & 40.37 \\

\cellcolor{lightblue}/w ReNIO
    & \cellcolor{lightblue}54.72
    & \cellcolor{lightblue}\textbf{43.06}
    & \cellcolor{lightblue}28.33
    & \cellcolor{lightblue}42.04 \\

OPD(4096) & 56.11 & 42.20 & 25.83 & 41.38 \\
\cellcolor{lightblue}/w ReNIO
    & \cellcolor{lightblue}\textbf{56.39}
    & \cellcolor{lightblue}41.94
    & \cellcolor{lightblue}\textbf{28.61}
    & \cellcolor{lightblue}\textbf{42.31} \\
\bottomrule
\end{tabular}
\caption{Effect of SGO prefix length on Qwen3-1.7B OPSD and OPD mathematical reasoning performance. Scores are Avg@$12$ over AIME24, AIME25, and HMMT25. 1024 and 4096 denote the maximum training-prefix length; 4096 can almost cover the final answer of an SGO in our setting.}
\label{tab:prefix-length-opsd}
\end{table}

\subsection{Ablation Studies}
\label{subsec:ablation}

All ablations in this subsection are conducted on Qwen3-1.7B under OPD, using Qwen3-8B as the teacher. We study the three components used to construct the final ReNIO weight: key-token selection, log-ratio clipping, and batch-wise weight normalization. Detailed hyperparameter ablation studies are provided in Appendix~\ref{app:additional-ablation-studies}.

\begin{table}[t]
\centering
\footnotesize
\setlength{\tabcolsep}{1.8pt}
\renewcommand{\arraystretch}{1.08}
\begin{tabular}{@{}lcccrrrr@{}}
\toprule
Method & Key & Clip & Norm & A24 & A25 & H25 & Avg \\
\midrule
OPD & No & No & No & 54.44 & 39.16 & 27.50 & 40.37 \\
\midrule
\cellcolor{lightblue}ReNIO
  & \cellcolor{lightblue}Yes
  & \cellcolor{lightblue}Yes
  & \cellcolor{lightblue}Yes
  & \cellcolor{lightblue}54.72
  & \cellcolor{lightblue}\textbf{43.06}
  & \cellcolor{lightblue}\textbf{28.33}
  & \cellcolor{lightblue}\textbf{42.04} \\
w/o clipping & Yes & No & Yes & 53.61 & 42.22 & 24.44 & 40.09 \\
w/o threshold & No & Yes & Yes & \textbf{55.28} & 41.94 & 25.28 & 40.83 \\
w/o batch norm & Yes & Yes & No & 51.94 & 41.39 & 25.28 & 39.54 \\

\bottomrule
\end{tabular}
\caption{Component ablation of ReNIO under OPD on Qwen3-1.7B mathematical reasoning. ``Key'' denotes threshold-based key-token selection, ``Clip'' denotes log-ratio clipping, and ``Norm'' denotes batch-wise weight normalization. Scores are Avg@$12$ on AIME24, AIME25, and HMMT25.}
\label{tab:component-ablation}
\end{table}
Table~\ref{tab:component-ablation} shows that the full pipeline gives the best average score, improving OPD from 40.37 to 42.04. Removing clipping reduces the average to 40.09, indicating that unbounded token ratios can make the sample weight unstable.

Key-token selection is also important. As shown in Figure~\ref{fig:log_ratio_dist}, most SGO tokens are already well matched by the student and teacher, so their ratios are close to 1. If these routine tokens are aggregated together with truly high-disagreement tokens, they dilute the signal from pivotal decisions. Consistently, removing threshold-based selection still improves over OPD but only reaches 40.83, below the full ReNIO result.

Batch-wise normalization has the largest effect among the ablations. Without it, the average drops to 39.54. This is expected because ReNIO's sample weights directly scale the distillation loss and therefore the gradient magnitude; without normalization, the effective objective scale can vary across batches and destabilize training.

\section{Related Work}
Classical knowledge distillation transfers teacher behavior to a student on fixed data or teacher-generated targets~\cite{hinton2015distilling}, but this offline setting can suffer from exposure bias in LLM post-training. On-policy distillation reduces this mismatch by sampling trajectories from the student policy and applying teacher supervision on the visited prefixes~\cite{agarwal2024onpolicy, GKD, ThinkingMachineOPD, minillm}. Recent OPD studies cover self-distillation \cite{OPSD, Shenfeld2026SelfDistillationEC, Hubotter2026ReinforcementLV, Kujanp2024EfficientKI, Ye2026OnPolicyCD, Stein2026GATESSU}, mechanism analysis \cite{whyOPSDNotWork, zhang2026prefixopd, li2026rethinkingopd, Yang2026LearningBT}, training stabilization \cite{relaxedOPD}, and better use of student-generated outputs \cite{Xu2024SpeculativeKD, kodistillm, ko2025distillm2, Lyu2025FromCT, Peng2025AdaSwitchBE}. Closest to our work, some methods reweight tokens or samples in OPD \cite{RLVR, scope, hou2026uniopd}, but their weighting or calibration signals require knowing whether an SGO reaches the correct answer. ReNIO instead computes sample weights from prefix-conditioned student and teacher probabilities, preserving OPD's short-prefix advantage while emphasizing likely negative trajectories.

\section{Conclusion}

This paper shows that incorrect SGOs can provide especially useful supervision for on-policy distillation, motivating a training strategy that emphasizes negative trajectories without observing final-answer correctness. We propose ReNIO, a prefix-computable reweighting method that identifies pivotal tokens through student-to-teacher log ratios and aggregates them into normalized sample weights. Because ReNIO does not require the answer of an SGO for weighting, it preserves OPD's advantage over RL by allowing training on SGO prefixes. Across mathematical reasoning and code generation tasks, ReNIO improves both OPD and OPSD while preserving short-prefix training.

\section*{Limitations}

We evaluate ReNIO across multiple model families and task domains, showing consistent gains on mathematical reasoning and code generation. However, due to hardware constraints, we have not verified its effectiveness on larger-scale models. 

\bibliography{custom}

\begin{thebibliography}{40}
\providecommand{\natexlab}[1]{#1}

\bibitem[{Agarwal et~al.(2024)Agarwal, Vieillard, Stanczyk, Ramos, Geist, and Bachem}]{agarwal2024onpolicy}
Rishabh Agarwal, Nino Vieillard, Piotr Stanczyk, Sabela Ramos, Matthieu Geist, and Olivier Bachem. 2024.
\newblock \href {https://openreview.net/forum?id=3zKtaqxLhW} {On-policy distillation of language models: Learning from self-generated mistakes}.
\newblock In \emph{Proceedings of the 12th International Conference on Learning Representations}.

\bibitem[{Agarwal et~al.(2023)Agarwal, Vieillard, Zhou, Stańczyk, Ramos, Geist, and Bachem}]{GKD}
Rishabh Agarwal, Nino Vieillard, Yongchao Zhou, Piotr Stańczyk, Sabela Ramos, Matthieu Geist, and Olivier Bachem. 2023.
\newblock \href {https://api.semanticscholar.org/CorpusID:263610088} {On-policy distillation of language models: Learning from self-generated mistakes}.
\newblock In \emph{International Conference on Learning Representations}.

\bibitem[{Bai et~al.(2025)Bai, Bao, Chen, Chen, Chen, Chen, Chen, Chen, Chen, Chen, Cui, Ding, xiao Dong, Du, Du, Du, Du, Fan, Feng, Fu, Gao, Gao, Gao, Gao, Gu, Guan, Guo, Guo, Hu, Hao, He, He, He, Hong, Hu, Hu, Huang, Huang, Huang, Jiang, Jiang, Jin, Kang, Lai, Li, Li, Li, Li, Li, Li, Li, Li, Li, Lin, Lin, Lin, Liu, Liu, Liu, Liu, Liu, Liu, Liu, Liu, Liu, Liu, Liu, Liu, Liu, Liu, Liu, Lu, Lu, Ma, Ma, Ma, Mao, Mei, Men, Miao, Pan, Peng, Qin, Qu, Shang, Shi, Shi, Song, Su, Su, Sun, Sung, Tang, Tao, Teng, Wang, Wang, Wang, Wang, Wang, Wang, Wang, Wang, Wang, Wang, Wang, Wang, Wang, Wang, Wang, Wang, Wang, Wei, Wei, Wu, Wu, Wu, Xiao, Xie, Xiong, Xu, Xu, Xu, Xu, Xu, Xu, Xu, Xu, Xu, Xu, Yan, Yan, Yang, Yang, Yang, Yang, Yang, Yao, Yao, guang Ye, Ye, Yin, Yu, Yuan, Yuan, Yuan, Zhan, Zhang, Zhang, Zhang, Zhang, Zhang, Zhang, Zhang, Zhang, Zhang, Zhang, Zhang, Zhao, Zhao, Zheng, Zheng, Zhou, Zhou, Zhou, Zhu, Zhuang, and Zu}]{kimik2}
Kimi Team~Yifan Bai, Yiping Bao, Guanduo Chen, Jiahao Chen, Ningxin Chen, Ruijue Chen, Yanru Chen, Yuankun Chen, Yutian Chen, Zhuofu Chen, Jialei Cui, Haochen Ding, Meng xiao Dong, Angang Du, Chenzhuang Du, Dikang Du, Yulun Du, Yu~Fan, Yichen Feng, and 149 others. 2025.
\newblock \href {https://api.semanticscholar.org/CorpusID:280323540} {Kimi k2: Open agentic intelligence}.

\bibitem[{Chen et~al.(2026)Chen, Ruan, Dan, Wang, Yan, Wu, Zhang, Chen, Zhou, He, Qi, Li, Guo, Shi, and Zhang}]{chen2026surveyinductivereasoninglarge}
Kedi Chen, Dezhao Ruan, Yuhao Dan, Yaoting Wang, Siyu Yan, Xuecheng Wu, Yinqi Zhang, Qin Chen, Jie Zhou, Liang He, Biqing Qi, Linyang Li, Qipeng Guo, Xiaoming Shi, and Wei Zhang. 2026.
\newblock \href {https://arxiv.org/abs/2510.10182} {A survey of inductive reasoning for large language models}.
\newblock \emph{Preprint}, arXiv:2510.10182.

\bibitem[{DeepSeek-AI(2025)}]{ds_distill_qwen}
DeepSeek-AI. 2025.
\newblock \href {https://arxiv.org/abs/2501.12948} {Deepseek-r1: Incentivizing reasoning capability in llms via reinforcement learning}.
\newblock \emph{Preprint}, arXiv:2501.12948.

\bibitem[{DeepSeek-AI et~al.(2025)DeepSeek-AI, Guo, Yang, Zhang, Song, Zhang, Xu, Zhu, Ma, Wang, Bi, Zhang, Yu, Wu, Wu, Gou, Shao, Li, Gao, Liu, Xue, Wang, Wu, Feng, Lu, Zhao, Deng, Zhang, Ruan, Dai, Chen, Ji, Li, Lin, Dai, Luo, Hao, Chen, Li, Zhang, Bao, Xu, Wang, Ding, Xin, Gao, Qu, Li, Guo, Li, Wang, Chen, Yuan, Qiu, Li, Cai, Ni, Liang, Chen, Dong, Hu, Gao, Guan, Huang, Yu, Wang, Zhang, Zhao, Wang, Zhang, Xu, Xia, Zhang, Zhang, Tang, Li, Wang, Li, Tian, Huang, Zhang, Wang, Chen, Du, Ge, Zhang, Pan, Wang, Chen, Jin, Chen, Lu, Zhou, Chen, Ye, Wang, Yu, Zhou, Pan, Li, Zhou, Wu, Yun, Pei, Sun, Wang, Zeng, Zhao, Liu, Liang, Gao, Yu, Zhang, Xiao, An, Liu, Wang, Chen, Nie, Cheng, Liu, Xie, Liu, Yang, Li, Su, Lin, Li, Jin, Shen, Chen, Sun, Wang, Song, Zhou, Wang, Shan, Li, Wang, Wei, Zhang, Xu, Li, Zhao, Sun, Wang, Yu, Zhang, Shi, Xiong, He, Piao, Wang, Tan, Ma, Liu, Guo, Ou, Wang, Gong, Zou, He, Xiong, Luo, mei You, Liu, Zhou, Zhu, Huang, Li, Zheng, Zhu, Ma, Tang, Zha, Yan, Ren, Ren, Sha, Fu, Xu, Xie, guo Zhang,
  Hao, Ma, Yan, Wu, Gu, Zhu, Liu, Li, Xie, Song, Pan, Huang, Xu, Zhang, and Zhang}]{DSR1}
DeepSeek-AI, Daya Guo, Dejian Yang, Haowei Zhang, Jun-Mei Song, Ruoyu Zhang, Runxin Xu, Qihao Zhu, Shirong Ma, Peiyi Wang, Xiaoling Bi, Xiaokang Zhang, Xingkai Yu, Yu~Wu, Z.~F. Wu, Zhibin Gou, Zhihong Shao, Zhuoshu Li, Ziyi Gao, and 179 others. 2025.
\newblock \href {https://api.semanticscholar.org/CorpusID:275789950} {Deepseek-r1 incentivizes reasoning in llms through reinforcement learning}.
\newblock \emph{Nature}, 645:633 -- 638.

\bibitem[{Dekoninck et~al.(2026)Dekoninck, Jovanović, Gehrunger, Rögnvaldsson, Petrov, Sun, and Vechev}]{hmmt25}
Jasper Dekoninck, Nikola Jovanović, Tim Gehrunger, Kári Rögnvaldsson, Ivo Petrov, Chenhao Sun, and Martin Vechev. 2026.
\newblock \href {https://arxiv.org/abs/2605.00674} {Beyond benchmarks: Matharena as an evaluation platform for mathematics with llms}.

\bibitem[{Gu et~al.(2023)Gu, Dong, Wei, and Huang}]{minillm}
Yuxian Gu, Li~Dong, Furu Wei, and Minlie Huang. 2023.
\newblock \href {https://api.semanticscholar.org/CorpusID:259164722} {Minillm: On-policy distillation of large language models}.

\bibitem[{Guha et~al.(2025)Guha, Marten, Keh, Raoof, Smyrnis, Bansal, Nezhurina, Mercat, Vu, Sprague, Suvarna, Feuer, Chen, Khan, Frankel, Grover, Choi, Muennighoff, Su, Zhao, Yang, Pimpalgaonkar, Sharma, Ji, Deng, Pratt, Ramanujan, Saad-Falcon, Li, Dave, Albalak, Arora, Wulfe, Hegde, Durrett, Oh, Bansal, Gabriel, Grover, Chang, Shankar, Gokaslan, Merrill, Hashimoto, Choi, Jitsev, Heckel, Sathiamoorthy, Dimakis, and Schmidt}]{Openthoughts}
Etash~Kumar Guha, Ryan Marten, Sedrick~Scott Keh, Negin Raoof, Georgios Smyrnis, Hritik Bansal, Marianna Nezhurina, Jean-Pierre Mercat, Trung Vu, Zayne Sprague, Ashima Suvarna, Ben Feuer, Liangyu Chen, Zaid Khan, Eric Frankel, Sachin Grover, Caroline Choi, Niklas Muennighoff, Shiye Su, and 31 others. 2025.
\newblock \href {https://api.semanticscholar.org/CorpusID:279154475} {Openthoughts: Data recipes for reasoning models}.
\newblock \emph{ArXiv}, abs/2506.04178.

\bibitem[{Hinton et~al.(2015)Hinton, Vinyals, and Dean}]{hinton2015distilling}
Geoffrey Hinton, Oriol Vinyals, and Jeff Dean. 2015.
\newblock \href {https://arxiv.org/abs/1503.02531} {Distilling the knowledge in a neural network}.
\newblock In \emph{NIPS Deep Learning and Representation Learning Workshop}.

\bibitem[{Hou et~al.(2026)Hou, Peng, Wang, Ruan, Zhang, Zhou, Gao, Chen, Wang, Yang, Zhang, Tian, Hu, Yang, Wu, and Fan}]{hou2026uniopd}
Wenjin Hou, Shangpin Peng, Weinong Wang, Zheng Ruan, Yue Zhang, Zhenglin Zhou, Mingqi Gao, Yifei Chen, Kaiqi Wang, Hongming Yang, Chengquan Zhang, Zhuotao Tian, Han Hu, Yi~Yang, Fei Wu, and Hehe Fan. 2026.
\newblock \href {https://arxiv.org/abs/2605.03677} {Uni-opd: Unifying on-policy distillation with a dual-perspective recipe}.
\newblock \emph{Preprint}, arXiv:2605.03677.

\bibitem[{Hubotter et~al.(2026)Hubotter, Lubeck, Behric, Baumann, Bagatella, Marta, Hakimi, Shenfeld, Buening, Guestrin, and Krause}]{Hubotter2026ReinforcementLV}
Jonas Hubotter, Frederike Lubeck, Lejs~Deen Behric, Anton Baumann, Marco Bagatella, Daniel Marta, Ido Hakimi, Idan Shenfeld, Thomas~Kleine Buening, Carlos Guestrin, and Andreas Krause. 2026.
\newblock \href {https://api.semanticscholar.org/CorpusID:285102353} {Reinforcement learning via self-distillation}.
\newblock \emph{ArXiv}, abs/2601.20802.

\bibitem[{Ke et~al.(2026)Ke, Jiao, Ming, Nguyen, Xu, Long, Li, Qin, Wang, Savarese, Xiong, and Joty}]{ke2026surveyfrontiersllmreasoning}
Zixuan Ke, Fangkai Jiao, Yifei Ming, Xuan-Phi Nguyen, Austin Xu, Do~Xuan Long, Minzhi Li, Chengwei Qin, Peifeng Wang, Silvio Savarese, Caiming Xiong, and Shafiq Joty. 2026.
\newblock \href {https://arxiv.org/abs/2504.09037} {A survey of frontiers in llm reasoning: Inference scaling, learning to reason, and agentic systems}.
\newblock \emph{Preprint}, arXiv:2504.09037.

\bibitem[{Kim et~al.(2026{\natexlab{a}})Kim, Luo, Kim, Lee, Kim, Jeon, Li, and Yang}]{whyOPSDNotWork}
Jeonghye Kim, Xufang Luo, Minbeom Kim, Sangmook Lee, Dohyung Kim, Jiwon Jeon, Dongsheng Li, and Yuqing Yang. 2026{\natexlab{a}}.
\newblock \href {https://api.semanticscholar.org/CorpusID:286776340} {Why does self-distillation (sometimes) degrade the reasoning capability of llms?}

\bibitem[{Kim et~al.(2026{\natexlab{b}})Kim, Luo, Kim, Lee, Li, and Yang}]{Kim2026UnderstandingRI}
Jeonghye Kim, Xufang Luo, Minbeom Kim, Sangmook Lee, Dongsheng Li, and Yuqing Yang. 2026{\natexlab{b}}.
\newblock \href {https://api.semanticscholar.org/CorpusID:286572576} {Understanding reasoning in llms through strategic information allocation under uncertainty}.

\bibitem[{Ko et~al.(2026)Ko, Abdali, Kim, Chen, and Cameron}]{relaxedOPD}
Jongwoo Ko, Sara Abdali, Young~Jin Kim, Tianyi Chen, and Pashmina Cameron. 2026.
\newblock \href {https://api.semanticscholar.org/CorpusID:286489350} {Scaling reasoning efficiently via relaxed on-policy distillation}.

\bibitem[{Ko et~al.(2025)Ko, Chen, Kim, Ding, Liang, Zharkov, and Yun}]{ko2025distillm2}
Jongwoo Ko, Tianyi Chen, Sungnyun Kim, Tianyu Ding, Luming Liang, Ilya Zharkov, and Se-Young Yun. 2025.
\newblock Distillm-2: A contrastive approach boosts the distillation of llms.
\newblock \emph{arXiv preprint arXiv:2503.07067}.

\bibitem[{Ko et~al.()Ko, Kim, Chen, and Yun}]{kodistillm}
Jongwoo Ko, Sungnyun Kim, Tianyi Chen, and Se-Young Yun.
\newblock Distillm: Towards streamlined distillation for large language models.
\newblock In \emph{Forty-first International Conference on Machine Learning}.

\bibitem[{Kujanp{\"a}{\"a} et~al.(2024)Kujanp{\"a}{\"a}, Marttinen, Valpola, and Ilin}]{Kujanp2024EfficientKI}
Kalle Kujanp{\"a}{\"a}, Pekka Marttinen, Harri Valpola, and Alexander Ilin. 2024.
\newblock \href {https://api.semanticscholar.org/CorpusID:274859956} {Efficient knowledge injection in llms via self-distillation}.
\newblock \emph{Trans. Mach. Learn. Res.}, 2025.

\bibitem[{Li et~al.(2026)Li, Zuo, He, Zhang, Xiao, Qian, Yu, Gao, Yang, Liu, and Ding}]{li2026rethinkingopd}
Yaxuan Li, Yuxin Zuo, Bingxiang He, Jinqian Zhang, Chaojun Xiao, Cheng Qian, Tianyu Yu, Huan-ang Gao, Wenkai Yang, Zhiyuan Liu, and Ning Ding. 2026.
\newblock \href {https://arxiv.org/abs/2604.13016} {Rethinking on-policy distillation of large language models: Phenomenology, mechanism, and recipe}.
\newblock \emph{arXiv preprint arXiv:2604.13016}.

\bibitem[{Liu et~al.(2023)Liu, Xia, Wang, and Zhang}]{evalplus}
Jiawei Liu, Chunqiu~Steven Xia, Yuyao Wang, and Lingming Zhang. 2023.
\newblock \href {https://openreview.net/forum?id=1qvx610Cu7} {Is your code generated by chat{GPT} really correct? rigorous evaluation of large language models for code generation}.
\newblock In \emph{Thirty-seventh Conference on Neural Information Processing Systems}.

\bibitem[{Loshchilov and Hutter(2019)}]{AdamW}
Ilya Loshchilov and Frank Hutter. 2019.
\newblock \href {https://openreview.net/forum?id=Bkg6RiCqY7} {Decoupled weight decay regularization}.
\newblock In \emph{International Conference on Learning Representations}.

\bibitem[{Lu and Lab(2025)}]{ThinkingMachineOPD}
Kevin Lu and Thinking~Machines Lab. 2025.
\newblock \href {https://doi.org/10.64434/tml.20251026} {On-policy distillation}.
\newblock \emph{Thinking Machines Lab: Connectionism}.
\newblock Https://thinkingmachines.ai/blog/on-policy-distillation.

\bibitem[{Lyu et~al.(2025)Lyu, Wang, Huang, and Xu}]{Lyu2025FromCT}
Yuanjie Lyu, Chengyu Wang, Jun Huang, and Tong Xu. 2025.
\newblock \href {https://api.semanticscholar.org/CorpusID:281393943} {From correction to mastery: Reinforced distillation of large language model agents}.
\newblock \emph{ArXiv}, abs/2509.14257.

\bibitem[{Peng et~al.(2025)Peng, Wang, Cai, Li, Zhang, Wang, Yin, and Zhao}]{Peng2025AdaSwitchBE}
Jingyu Peng, Maolin Wang, Hengyi Cai, Yuchen Li, Kai Zhang, Shuaiqiang Wang, Dawei Yin, and Xiangyu Zhao. 2025.
\newblock \href {https://api.semanticscholar.org/CorpusID:286579337} {Adaswitch: Balancing exploration and guidance in knowledge distillation via adaptive switching}.

\bibitem[{Shao et~al.(2024)Shao, Wang, Zhu, Xu, Song, Zhang, Li, Wu, and Guo}]{deepseek-math}
Zhihong Shao, Peiyi Wang, Qihao Zhu, Runxin Xu, Junxiao Song, Mingchuan Zhang, Y.K. Li, Y.~Wu, and Daya Guo. 2024.
\newblock \href {https://arxiv.org/abs/2402.03300} {Deepseekmath: Pushing the limits of mathematical reasoning in open language models}.

\bibitem[{Shenfeld et~al.(2026)Shenfeld, Damani, H{\"u}botter, and Agrawal}]{Shenfeld2026SelfDistillationEC}
Idan Shenfeld, Mehul Damani, Jonas H{\"u}botter, and Pulkit Agrawal. 2026.
\newblock \href {https://api.semanticscholar.org/CorpusID:285071839} {Self-distillation enables continual learning}.
\newblock \emph{ArXiv}, abs/2601.19897.

\bibitem[{Stein et~al.(2026)Stein, Huang, and Goldstein}]{Stein2026GATESSU}
Alex Stein, Furong Huang, and Tom Goldstein. 2026.
\newblock \href {https://api.semanticscholar.org/CorpusID:286001285} {Gates: Self-distillation under privileged context with consensus gating}.
\newblock \emph{ArXiv}, abs/2602.20574.

\bibitem[{Wen et~al.(2025)Wen, Liu, Zheng, Xu, Ye, Wu, Liang, Wang, Li, Miao, Bian, and Yang}]{RLVR}
Xumeng Wen, Zihan Liu, Shun Zheng, Zhijian Xu, Shengyu Ye, Zhirong Wu, Xiao Liang, Yang Wang, Junjie Li, Ziming Miao, Jiang Bian, and Mao Yang. 2025.
\newblock \href {https://api.semanticscholar.org/CorpusID:279410727} {Reinforcement learning with verifiable rewards implicitly incentivizes correct reasoning in base llms}.
\newblock \emph{ArXiv}, abs/2506.14245.

\bibitem[{Xu et~al.(2024)Xu, Han, Wang, Le, Madeka, Li, Wang, Agarwal, Lee, and Pfister}]{Xu2024SpeculativeKD}
Wenda Xu, Rujun Han, Zifeng Wang, Long~T. Le, Dhruv Madeka, Lei Li, William~Yang Wang, Rishabh Agarwal, Chen-Yu Lee, and Tomas Pfister. 2024.
\newblock \href {https://api.semanticscholar.org/CorpusID:273350649} {Speculative knowledge distillation: Bridging the teacher-student gap through interleaved sampling}.
\newblock \emph{ArXiv}, abs/2410.11325.

\bibitem[{Yang et~al.(2025)Yang, Li, Yang, Zhang, Hui, Zheng, Yu, Gao, Huang, Lv, Zheng, Liu, Zhou, Huang, Hu, Ge, Wei, Lin, Tang, Yang, Tu, Zhang, Yang, Yang, Zhou, Zhou, Lin, Dang, Bao, Yang, Yu, Deng, Li, Xue, Li, Zhang, Wang, Zhu, Men, Gao, Liu, Luo, Li, Tang, Yin, Ren, Wang, Zhang, Ren, Fan, Su, Zhang, Zhang, Wan, Liu, Wang, Cui, Zhang, Zhou, and Qiu}]{Qwen3}
An~Yang, Anfeng Li, Baosong Yang, Beichen Zhang, Binyuan Hui, Bo~Zheng, Bowen Yu, Chang Gao, Chengen Huang, Chenxu Lv, Chujie Zheng, Dayiheng Liu, Fan Zhou, Fei Huang, Feng Hu, Hao Ge, Haoran Wei, Huan Lin, Jialong Tang, and 41 others. 2025.
\newblock \href {https://api.semanticscholar.org/CorpusID:278602855} {Qwen3 technical report}.

\bibitem[{Yang et~al.(2026)Yang, Liu, Xie, Yang, Yang, and Lin}]{Yang2026LearningBT}
Wenkai Yang, Weijie Liu, Ruobing Xie, Kai Yang, Saiyong Yang, and Yankai Lin. 2026.
\newblock \href {https://api.semanticscholar.org/CorpusID:285540530} {Learning beyond teacher: Generalized on-policy distillation with reward extrapolation}.
\newblock \emph{ArXiv}, abs/2602.12125.

\bibitem[{Ye et~al.(2026)Ye, Dong, Wu, Huang, and Wei}]{Ye2026OnPolicyCD}
Tianzhu Ye, Li~Dong, Xun Wu, Shaohan Huang, and Furu Wei. 2026.
\newblock \href {https://api.semanticscholar.org/CorpusID:285540636} {On-policy context distillation for language models}.
\newblock \emph{ArXiv}, abs/2602.12275.

\bibitem[{Yu et~al.(2025)Yu, Zhang, Zhu, Yuan, Zuo, Yue, Fan, Liu, Liu, Liu, Lin, Lin, Ma, Sheng, Tong, Zhang, Zhang, Zhang, Zhu, Zhu, Chen, Chen, Wang, Yu, Dai, Song, Wei, Zhou, Liu, Ma, Zhang, Yan, Qiao, Wu, and Wang}]{dapo}
Qiying Yu, Zheng Zhang, Ruofei Zhu, Yufeng Yuan, Xiaochen Zuo, Yu~Yue, Tiantian Fan, Gaohong Liu, Lingjun Liu, Xin Liu, Haibin Lin, Zhiqi Lin, Bole Ma, Guangming Sheng, Yuxuan Tong, Chi Zhang, Mofan Zhang, Wang Zhang, Hang Zhu, and 16 others. 2025.
\newblock \href {https://api.semanticscholar.org/CorpusID:277104124} {Dapo: An open-source llm reinforcement learning system at scale}.
\newblock \emph{ArXiv}, abs/2503.14476.

\bibitem[{Zhang et~al.(2026)Zhang, Yang, Janghorbani, Han, Ressler, Qian, Lyng, Batra, and Tillman}]{zhang2026prefixopd}
Dan Zhang, Zhuohan Yang, Sepehr Janghorbani, Jiaqi Han, Andrew Ressler, Qi~Qian, Gabriel~D. Lyng, Sarthak~S. Batra, and Richard~E. Tillman. 2026.
\newblock Fast and effective on-policy distillation from reasoning prefixes.
\newblock \emph{Manuscript}.

\bibitem[{Zhang and Math-AI(2024)}]{aime24}
Yifan Zhang and Team Math-AI. 2024.
\newblock American invitational mathematics examination (aime) 2024.

\bibitem[{Zhang and Math-AI(2025)}]{aime25}
Yifan Zhang and Team Math-AI. 2025.
\newblock American invitational mathematics examination (aime) 2025.

\bibitem[{Zhao et~al.(2026)Zhao, Xie, Liu, Huang, Pang, Chen, and Grover}]{OPSD}
Siyan Zhao, Zhihui Xie, Mengchen Liu, Jing Huang, Guan Pang, Feiyu Chen, and Aditya Grover. 2026.
\newblock \href {https://api.semanticscholar.org/CorpusID:285050713} {Self-distilled reasoner: On-policy self-distillation for large language models}.
\newblock \emph{ArXiv}, abs/2601.18734.

\bibitem[{Zheng et~al.(2026)Zheng, Ma, Liang, Ruan, Fu, Lin, Zhu, Zeng, and Cai}]{scope}
Binbin Zheng, Xing Ma, Yiheng Liang, Jingqing Ruan, Xiaoliang Fu, Kepeng Lin, Benchang Zhu, Ke~Zeng, and Xunliang Cai. 2026.
\newblock \href {https://arxiv.org/abs/2604.10688} {Scope: Signal-calibrated on-policy distillation enhancement with dual-path adaptive weighting}.
\newblock \emph{Preprint}, arXiv:2604.10688.

\bibitem[{Zhu et~al.(2025)Zhu, Xia, Wei, Chen, Chen, and Meng}]{negImportant}
Xinyu Zhu, Mengzhou Xia, Zhepei Wei, Wei-Lin Chen, Danqi Chen, and Yu~Meng. 2025.
\newblock \href {https://api.semanticscholar.org/CorpusID:279075301} {The surprising effectiveness of negative reinforcement in llm reasoning}.
\newblock \emph{ArXiv}, abs/2506.01347.

\end{thebibliography}

\appendix
\section{Reverse-KL Gradient Interpretation of the Student--Teacher Ratio}
\label{app:rkl-gradient}

This appendix derives why the student-to-teacher log ratio used by ReNIO can be interpreted as a token-level gradient weight under reverse-KL distillation. Consider one fixed prefix $(x,y_{<t})$ and write the student and teacher next-token distributions as $p_S(v)$ and $p_T(v)$. The reverse-KL objective at this prefix is
\begin{equation}
\begin{aligned}
\mathcal{L}_{\mathrm{RKL}}
&= \mathrm{KL}(p_S \| p_T) \\
&= \sum_{v} p_S(v)\bigl(\log p_S(v)-\log p_T(v)\bigr).
\end{aligned}
\label{eq:app-rkl}
\end{equation}
Let $z_u$ be the student logit for token $u$, so that $p_S(v)=\operatorname{softmax}(z)_v$. Differentiating Eq.~\ref{eq:app-rkl} with respect to $z_u$ gives
\begin{equation}
\begin{aligned}
\frac{\partial \mathcal{L}_{\mathrm{RKL}}}{\partial z_u}
&= \sum_v
\frac{\partial p_S(v)}{\partial z_u}\\
&\bigl(\log p_S(v)-\log p_T(v)+1\bigr).
\end{aligned}
\label{eq:app-rkl-grad-start}
\end{equation}
Using the softmax derivative
$\partial p_S(v)/\partial z_u
= p_S(v)(\mathbb{I}[v=u]-p_S(u))$,
we obtain
\begin{equation}
\begin{aligned}
&\frac{\partial \mathcal{L}_{\mathrm{RKL}}}{\partial z_u}
= p_S(u)\bigl(\log p_S(u)-\log p_T(u)+1\bigr) \\
&\quad - p_S(u)\sum_v p_S(v)
\bigl(\log p_S(v)-\log p_T(v)+1\bigr).
\end{aligned}
\label{eq:app-rkl-grad}
\end{equation}
The second term is shared across all vocabulary tokens at the same prefix and acts as a distribution-level baseline. The token-specific part is therefore controlled by $\log p_S(u)-\log p_T(u)+1=1+\log \frac{p_S(u)}{p_T(u)}$.
Equivalently, the same result can be written in score-function form:
\begin{equation}
\begin{aligned}
\nabla_\theta \mathcal{L}_{\mathrm{RKL}}
&= \mathbb{E}_{v\sim p_S}
\bigl[
(\log p_S(v)-\log p_T(v)+1) \\
&\qquad\qquad\qquad\cdot \nabla_\theta \log p_S(v)
\bigr].
\end{aligned}
\label{eq:app-score-function}
\end{equation}
Because $\mathbb{E}_{v\sim p_S}[\nabla_\theta \log p_S(v)]=\nabla_\theta \sum_v p_S(v)=0$, the constant $+1$ is a removable baseline. Therefore,
\begin{equation}
\begin{aligned}
\nabla_\theta \mathcal{L}_{\mathrm{RKL}}
&= \mathbb{E}_{v\sim p_S}
\bigl[
(\log p_S(v)-\log p_T(v)) \\
&\qquad\qquad\qquad\cdot \nabla_\theta \log p_S(v)
\bigr],
\end{aligned}
\label{eq:app-effective-rkl-grad}
\end{equation}
so the effective token-level weight is exactly the student-to-teacher log ratio.

For an on-policy SGO, the observed token $y_t$ is sampled from the student distribution. Substituting $v=y_t$ into Eq.~\ref{eq:app-effective-rkl-grad} shows that the sampled token's effective reverse-KL gradient contribution is weighted by
\begin{equation}
\ell_t=\log \frac{p_S(y_t\mid x,y_{<t})}{p_T(y_t\mid x,y_{<t})}.
\end{equation}

Thus, tokens that the student assigns much higher probability than the teacher receive larger reverse-KL corrective emphasis. ReNIO uses this same log-ratio term as token-level evidence for identifying pivotal decisions and then aggregates the selected evidence into a normalized sample-level weight.

\section{ReNIO Algorithm}
\label{app:algorithm}

\begin{algorithm}[H]
\caption{\textbf{ReNIO} Algorithm for On-Policy Distillation}
\label{alg:weight}
\begin{algorithmic}[1]
\REQUIRE Student $S$, teacher $T$, on-policy batch $\mathcal{B}=\{(x_i,y_i)\}_{i=1}^{B}$, threshold $\tau$, clip bounds $\epsilon_{\min},\epsilon_{\max}$
\ENSURE Normalized sample weights $\{\hat{w}_i\}_{i=1}^{B}$
\FOR{each sample $(x_i, y_i) \in \mathcal{B}$}
    \STATE \textcolor{blue}{\textbf{Phase~I: Student-Teacher Log Ratio}}
    \FOR{each token position $t = 1, \ldots, |y_i|$}
        \STATE \textcolor{blue}{$\ell_t \leftarrow \log p_S(y_t \mid x_i, y_{<t}) - \log p_T(y_t \mid x_i, y_{<t})$}
    \ENDFOR
    \STATE \textcolor{red}{\textbf{Phase~II: Key Token Information Selection}}
    \STATE \textcolor{red}{$\mathcal{K}_i \leftarrow \emptyset$}
    \FOR{each token position $t = 1, \ldots, |y_i|$}
        \IF{$\ell_t > \tau$}
            \STATE \textcolor{red}{$\mathcal{K}_i \leftarrow \mathcal{K}_i \cup \{t\}$}
        \ENDIF
    \ENDFOR
    \STATE \textcolor{green!60!black}{\textbf{Phase~III: Token Information Aggregation}}
    \IF{$\mathcal{K}_i \neq \emptyset$}
        \STATE \textcolor{green!60!black}{$\bar{\ell}_i \leftarrow \frac{1}{|\mathcal{K}_i|}\sum_{t \in \mathcal{K}_i}\operatorname{clip}(\ell_t,\; \epsilon_{\min},\; \epsilon_{\max})$}
        \STATE \textcolor{green!60!black}{$w_i \leftarrow \exp(\bar{\ell}_i)$}
    \ELSE
        \STATE $w_i \leftarrow 1$ \hfill \COMMENT{no pivotal token found}
    \ENDIF
\ENDFOR
\STATE $\bar{w}_B \leftarrow \frac{1}{B}\sum_{i=1}^{B} w_i$
\FOR{$i = 1, \ldots, B$}
    \STATE $\hat{w}_i \leftarrow w_i \;/\; \bar{w}_B$
\ENDFOR
\RETURN $\{\hat{w}_i\}_{i=1}^{B}$
\end{algorithmic}
\end{algorithm}


\section{Detailed Experimental Setup\label{sec:Detailed Experimental Setup}}
Here we list the detailed hyperparameter values for each setting in Section \ref{section: Main Experiments}. We provide the training and evaluation configurations for GRPO, OPD, OPSD experiments in Tables \ref{tab:training-config-qwen3}, \ref{tab:training-config-DS-Distill-Qwen}, \ref{tab:Evaliation Details}. Following the settings in OPSD ~\cite{OPSD}, for Qwen3 series models' OPD and OPSD training, we apply Thinking-Mode-off student/Thinking-Mode-on teacher, and for their evaluation, we apply Thinking-Mode on.

All experiments are applied on 4 H200 GPUs with gradient checkpointing and Flash Attention 2. We utilize the AdamW optimizer\cite{AdamW} and bfloat16 precision for all training runs.And we apply full-vocabulary logit distillation for all OPD and OPSD experiments.

\begin{table}[H]
\centering
\footnotesize
\setlength{\tabcolsep}{8pt}
\renewcommand{\arraystretch}{1.12}
\begin{tabular}{@{}lccc@{}}
\toprule
Parameter & GRPO & OPD & OPSD \\
\midrule
Learning rate & 5e-6 & 5e-6 & 5e-6 \\
Effective batch size & 32 & 32 & 32 \\
\midrule
LoRA rank ($r$) & 64 & 64 & 64 \\
LoRA alpha ($\alpha$) & 128 & 128 & 128 \\
LoRA target modules & \multicolumn{3}{c@{}}{\makecell[c]{q proj, k proj, v proj, o proj,\\ gate proj, up proj, down proj}} \\
\midrule
Max completion length & 16000 & 1024 & 1024 \\
\midrule
\makecell[l]{Number of generations\\per prompt} & 8 & 1 & 1 \\
Sampling temperature & 1.2 & 1.1 & 1.1 \\
Training steps & 300 & 100 & 100 \\
\bottomrule
\end{tabular}
\caption{Training configuration for Qwen3 series models.}
\label{tab:training-config-qwen3}
\end{table}

\begin{table}[H]
\centering
\footnotesize
\setlength{\tabcolsep}{8pt}
\renewcommand{\arraystretch}{1.12}
\begin{tabular}{@{}lccc@{}}
\toprule
Parameter & GRPO & OPD & OPSD \\
\midrule
Learning rate & 1e-6 & 1e-6 & 1e-6 \\
Effective batch size & 32 & 32 & 32 \\
\midrule
LoRA rank ($r$) & 64 & 64 & 64 \\
LoRA alpha ($\alpha$) & 128 & 128 & 128 \\
LoRA target modules & \multicolumn{3}{c@{}}{\makecell[c]{q proj, k proj, v proj, o proj,\\ gate proj, up proj, down proj}} \\
\midrule
Max completion length & 16000 & 2048 & 2048 \\
\midrule
\makecell[l]{Number of generations\\per prompt} & 8 & 1 & 1 \\
Sampling temperature & 1.2 & 1.1 & 1.1 \\
Training steps & 300 & 100 & 100 \\
\bottomrule
\end{tabular}
\caption{Training configuration for DS-Distill-Qwen series models.}
\label{tab:training-config-DS-Distill-Qwen}
\end{table}

\begin{table}[H]
\centering
\footnotesize
\setlength{\tabcolsep}{8pt}
\renewcommand{\arraystretch}{1.12}
\begin{tabular}{@{}lcc@{}}
\toprule
Parameter & Math & Coding \\
\midrule
Max New Tokens & 38912 & 38912 \\
Thinking Mode & Enabled  & Enabled \\
Temperature & 1.0 & 1.0 \\
Top-p & 0.95 & 0.95 \\
Top-k & -1 & -1 \\
Min-p & 0.0 & 0.0 \\
Presence Penalty & 0.0 & 0.0 \\
Samples per Prompt & 12 & 4 \\
\bottomrule
\end{tabular}
\caption{Evaluation configuration}
\label{tab:Evaliation Details}
\end{table}

Additionally, for experiments conducted in Section~\ref{section:motivtion experiments},
We keep most training and evaluation settings the same as those in Tables~\ref{tab:training-config-qwen3} and~\ref{tab:Evaliation Details}, except for the maximum completion length. Since we need the final output to identify whether the SGO is correct, we must roll out the whole trajectory. For Qwen3-1.7B with Thinking Mode off, a generation length of 4096 is enough to generate the final answer; hence, we set the maximum generation length to 4096 for experiments in Section~\ref{section:motivtion experiments}.

\section{Additional Analyses and Discussions}
\subsection{Teacher Confidence on High-Weight Trajectories}
\label{subsec:train-inference-mismatch}

A natural concern for ReNIO is that it may overweight low-quality SGOs produced by train--inference mismatch. If such trajectories fall outside the teacher's reliable region, their supervision could be weak or misleading. We examine this issue on Qwen3-1.7B using teacher entropy as a proxy for confidence: lower entropy indicates sharper and more reliable token-level guidance.

\begin{figure}[t]
\centering
\begin{subfigure}[t]{0.7\columnwidth}
    \centering
    \includegraphics[width=\linewidth]{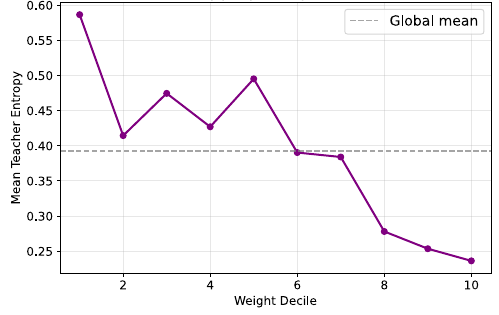}
    \caption{OPD}
    \label{fig:weight_entropy_opd}
\end{subfigure}
\hfill
\begin{subfigure}[t]{0.7\columnwidth}
    \centering
    \includegraphics[width=\linewidth]{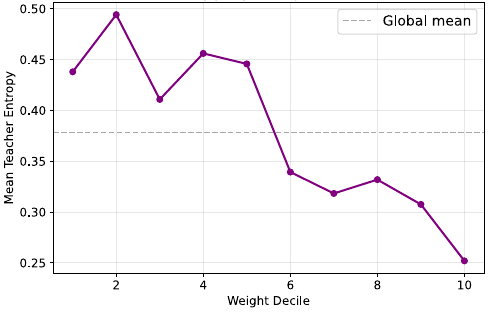}
    \caption{OPSD}
    \label{fig:weight_entropy_opsd}
\end{subfigure}
\caption{Relationship between ReNIO sample weight and teacher entropy on Qwen3-1.7B.}
\label{fig:weight_entropy}
\end{figure}


Figure~\ref{fig:weight_entropy} shows that higher ReNIO weights correlate with lower teacher entropy in both OPD and OPSD. This inverse relationship indicates that ReNIO does not primarily amplify trajectories for which the teacher distribution becomes uncertain. Instead, ReNIO assigns larger weights to SGOs where the student departs from the teacher but the teacher can still provide a sharp correction signal.

This pattern is plausible because current student models are already capable enough that an incorrect trajectory does not necessarily collapse into incoherent reasoning. As illustrated in Figure~\ref{fig:pipeline_b}, the student chooses the wrong pivotal token ``13'', yet the subsequent reasoning still follows a structured path conditioned on that mistaken branch. Such trajectories remain within a region where the teacher can recognize the local error with high confidence: the student assigns high probability to the sampled token, while the teacher assigns it low probability, producing a large student-to-teacher ratio and therefore a large ReNIO weight.

The opposite case explains the low-weight, high-entropy region in Figure~\ref{fig:weight_entropy}. If a student trajectory is far outside the teacher's reliable reasoning distribution, the teacher distribution becomes flatter and less decisive. In that case, the denominator in the student-to-teacher ratio is no longer sharply suppressed for a particular wrong token, so the ratio may become smaller rather than larger. Extremely erroneous trajectories therefore receive small ReNIO weights and are not strongly amplified. Overall, the negative correlation between ReNIO weight and teacher entropy supports the intended behavior of the method: \textbf{high weights correspond to student errors on which the teacher remains confident enough to give clear supervision. This further validates the Rationality of ReNIO.}

\subsection{Additional Discussion on ReNIO Weighting}
\label{app:additional-weighting-discussion}

To further examine the design of ReNIO, we compare it with two alternative weighting strategies under Qwen3-1.7B OPSD. The first variant uses the same student-to-teacher ratio signal as ReNIO but applies the resulting weights directly to token losses instead of aggregating them into a sample-level weight. The second variant keeps ReNIO's sample-level weighting form but reverses the information signal, using the teacher-to-student ratio instead of the student-to-teacher ratio.

\begin{table}[H]
\centering
\scriptsize
\setlength{\tabcolsep}{3.0pt}
\renewcommand{\arraystretch}{1.08}
\begin{tabular}{@{}lcccc@{}}
\toprule
Method & AIME24 & AIME25 & HMMT25 & Avg \\
\midrule
Base & 51.33 & 36.00 & 23.33 & 36.89 \\
OPSD & 53.06 & 41.94 & 27.50 & 40.83 \\
\midrule
\cellcolor{lightblue}/w ReNIO
    & \cellcolor{lightblue}\textbf{57.78}
    & \cellcolor{lightblue}\textbf{42.78}
    & \cellcolor{lightblue}27.78
    & \cellcolor{lightblue}\textbf{42.78} \\
/w T/S sample weighting & 55.28 & 40.00 & \textbf{29.44} & 41.57 \\
/w S/T token weighting & 54.72 & 40.28 & 25.83 & 40.28 \\
\bottomrule
\end{tabular}
\caption{Comparison of alternative weighting strategies under Qwen3-1.7B OPSD. Scores are Avg@$12$ on AIME24, AIME25, and HMMT25. S/T token weighting applies student-to-teacher ratio weights at the token level; T/S sample weighting keeps sample-level aggregation but uses the reverse teacher-to-student ratio.}
\label{tab:additional-weighting-discussion}
\end{table}

Table~\ref{tab:additional-weighting-discussion} shows that directly applying the S/T signal at the token level is ineffective: its average score is even lower than the OPSD baseline. This indicates that the S/T ratio should not be used to rescale every token loss independently; instead, it is more useful after being aggregated into a sample-level signal, as in ReNIO. The T/S sample-weighting variant improves over OPSD, but still underperforms ReNIO by 1.21 points. A likely explanation is that high T/S ratios emphasize tokens that the teacher assigns high probability but the student assigns low probability; such tokens may occur more often in correct or teacher-aligned SGOs, giving larger weights to trajectories that are not the main source of the negative-trajectory signal identified in Section~\ref{section:motivtion experiments}. This supports ReNIO's design choice of using S/T ratios to emphasize likely negative, high-information SGOs.

\subsection{Additional Ablation Studies}
\label{app:additional-ablation-studies}
\begin{figure*}[t]
\centering
\begin{subfigure}[b]{\textwidth}
    \centering
    \includegraphics[width=\linewidth]{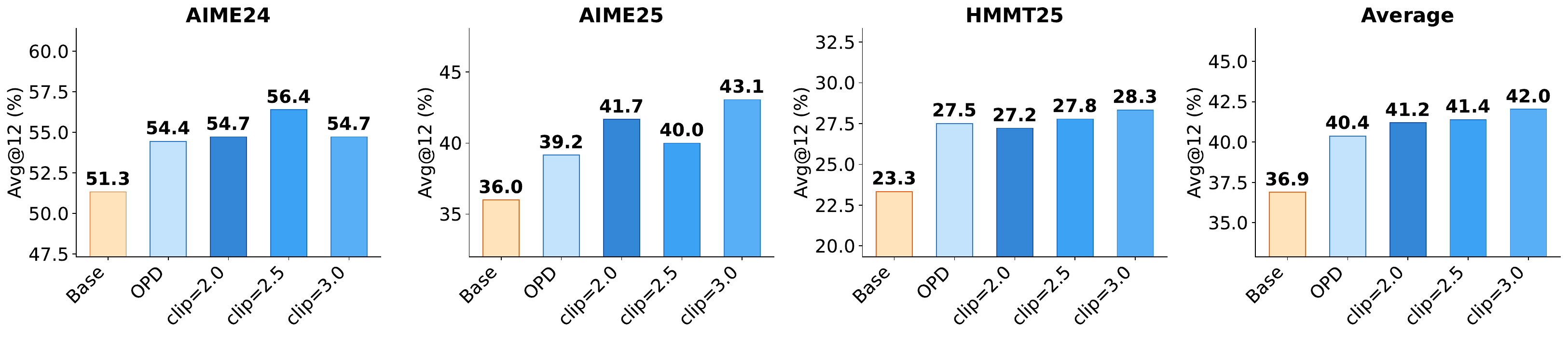}
    \caption{Effect of the clipping bound on OPD. The threshold is fixed to 0.8.}
    \label{fig:ablation_clip_full}
\end{subfigure}

\vspace{0.35em}

\begin{subfigure}[b]{\textwidth}
    \centering
    \includegraphics[width=\linewidth]{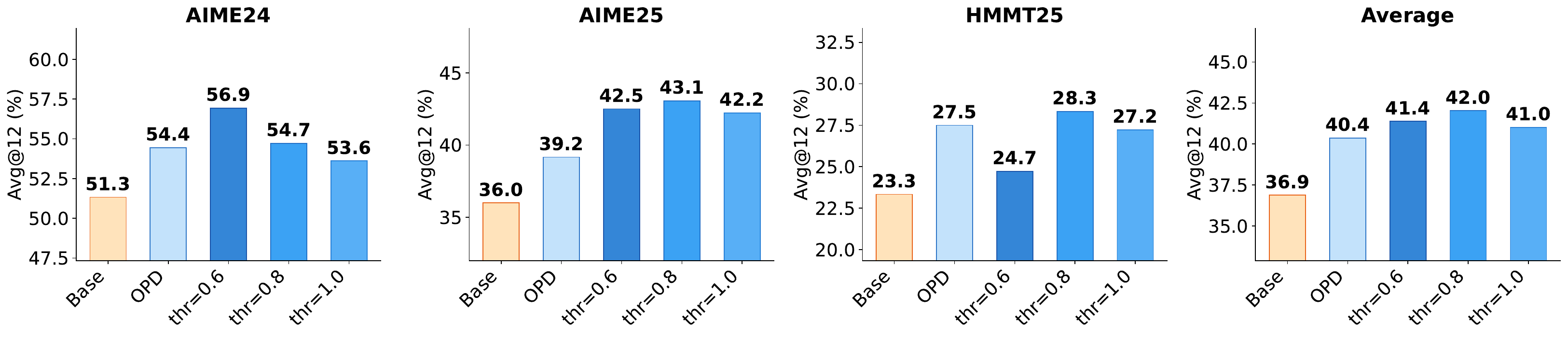}
    \caption{Effect of the key-token threshold on OPD. The clipping bound is fixed to 3.0.}
    \label{fig:ablation_thr_full}
\end{subfigure}
\caption{Per-benchmark hyperparameter ablation results for ReNIO under OPD on Qwen3-1.7B mathematical reasoning.}
\label{fig:ablation_hyperparams_full}
\end{figure*}

We vary the two main hyperparameters: the clipping bound and the key-token threshold on OPD on Qwen3-1.7B. Figure~\ref{fig:ablation_hyperparams_full} summarizes the per-benchmark results across the three mathematical reasoning benchmarks.

A small clipping bound suppresses useful disagreement, while a loose bound gives too much influence to extreme ratios; the best average is obtained at a clipping bound of 3.0. For key-token selection, a low threshold admits routine tokens and dilutes the signal, whereas a high threshold discards informative disagreements; the best average is obtained at a threshold of 0.8. Together, these results support the design of ReNIO: it emphasizes salient student--teacher disagreements while keeping sample weights numerically stable.

\section{Additional Experimental Results}
\label{app:additional-experimental-results}

\begin{figure*}[t]
\centering
\includegraphics[width=0.32\textwidth]{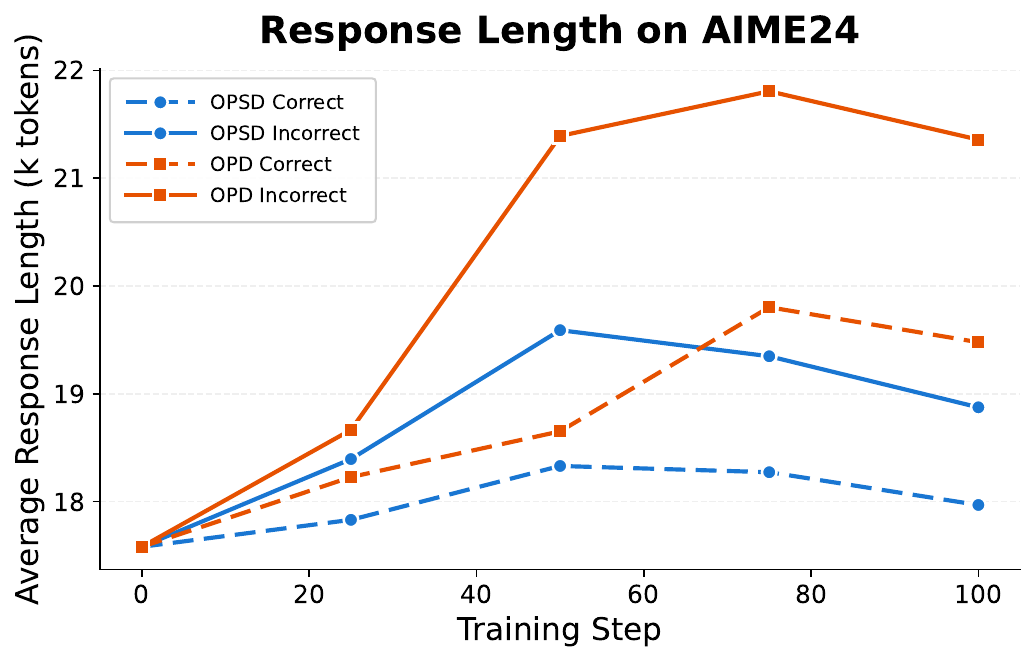}
\hfill
\includegraphics[width=0.32\textwidth]{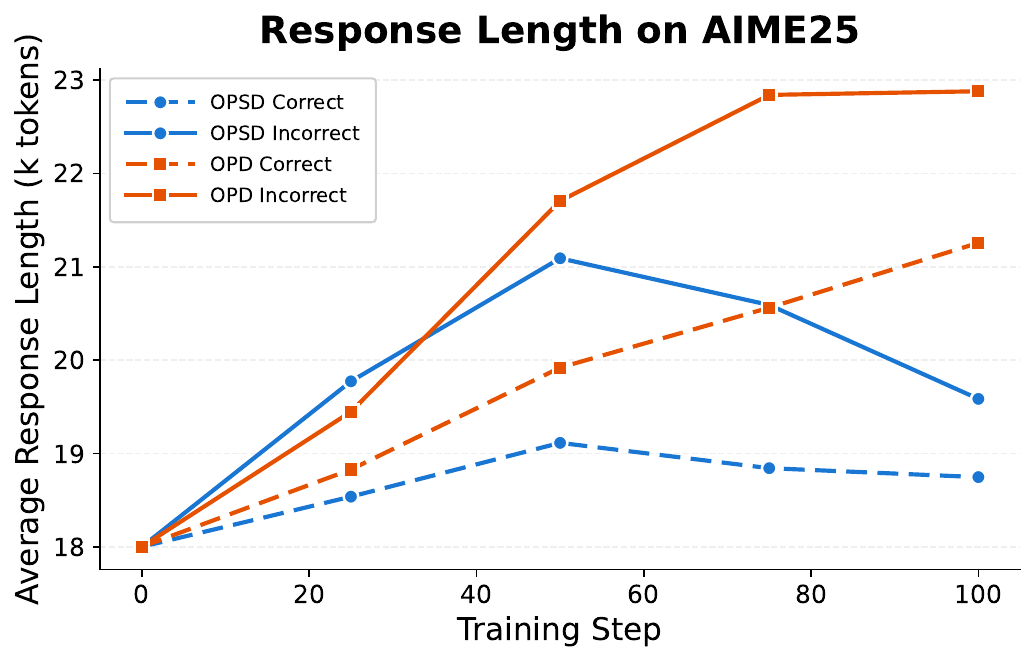}
\hfill
\includegraphics[width=0.32\textwidth]{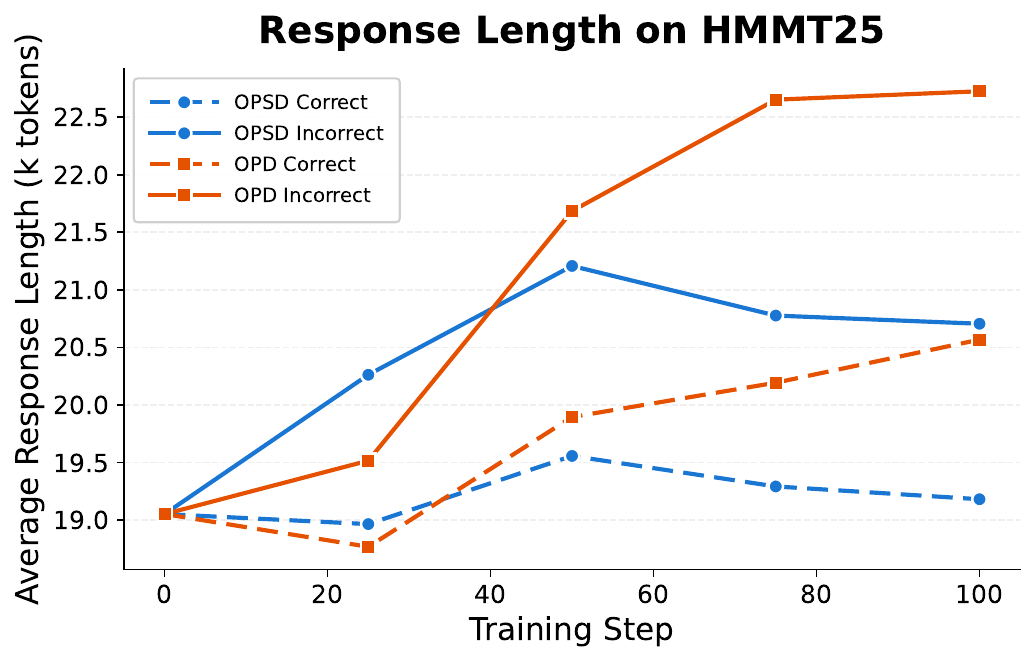}
\caption{Average response length of correct versus incorrect on-policy trajectories during training on AIME24 (left), AIME25 (middle), and HMMT25 (right). Incorrect trajectories are consistently longer, and the gap widens over training steps, indicating that incorrect trajectories accumulate more deviation tokens.}
\label{fig:gen_len_full}
\end{figure*}

\begin{figure*}[t]
\centering
\includegraphics[width=0.32\textwidth]{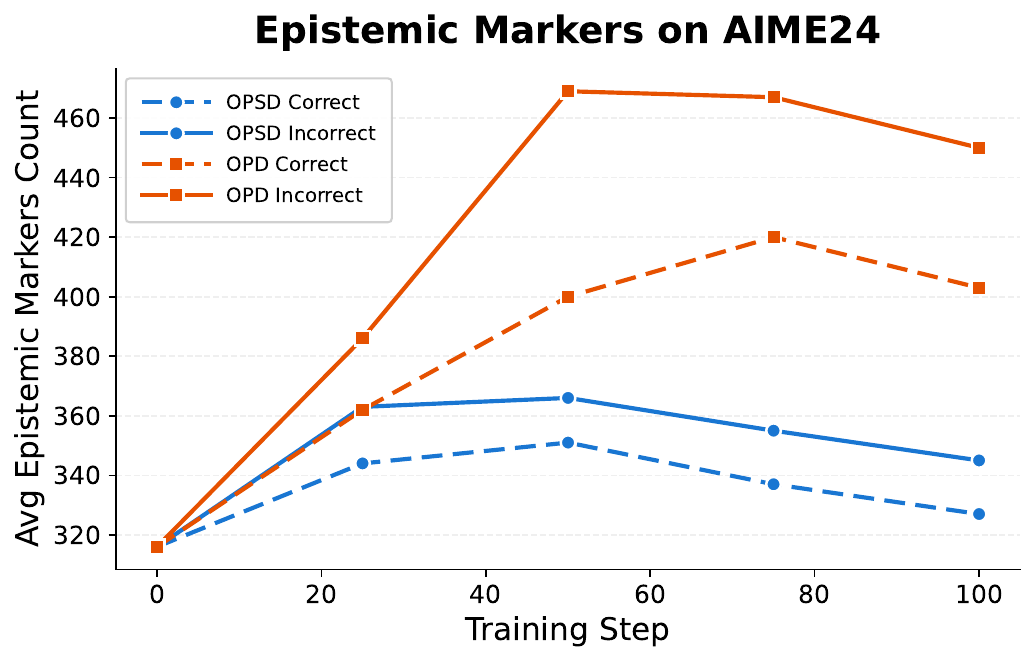}
\hfill
\includegraphics[width=0.32\textwidth]{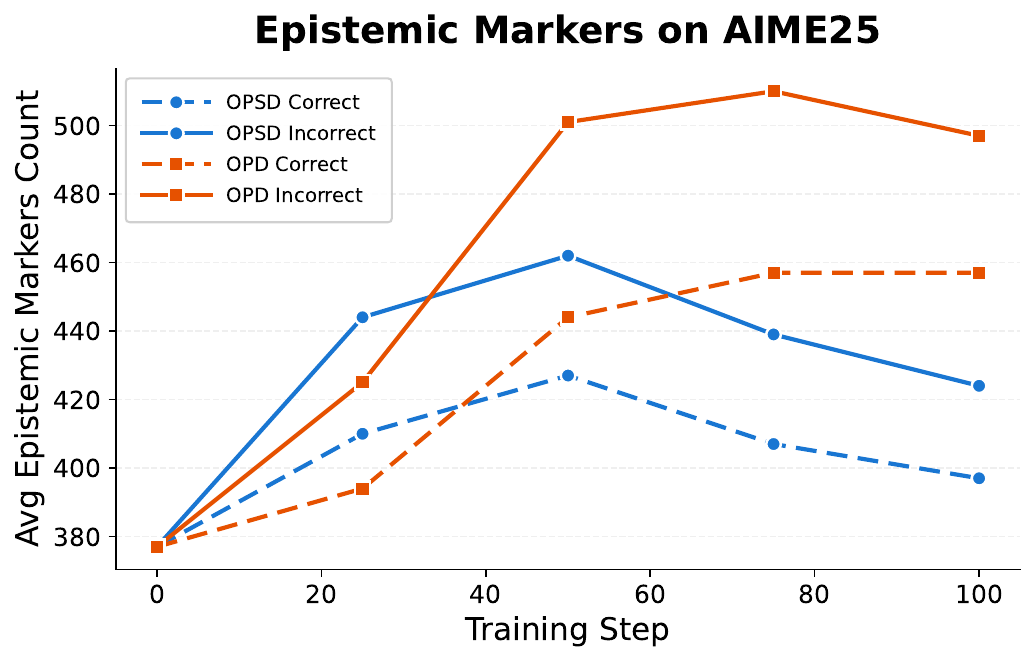}
\hfill
\includegraphics[width=0.32\textwidth]{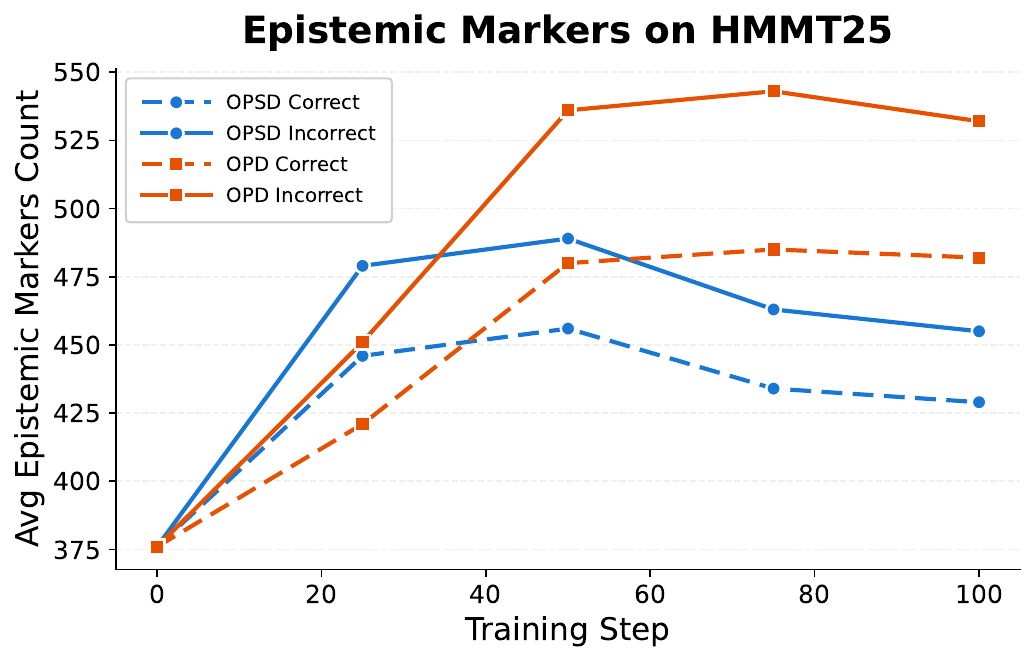}
\caption{Average count of Epistemic Markers of correct versus incorrect on-policy trajectories during training on AIME24 (left), AIME25 (middle), and HMMT25 (right). }
\label{fig:key_token_count_full}
\end{figure*}

Here we provide the full experimental results for Section~\ref{subsection:length experiments}, as shown in Figures~\ref{fig:gen_len_full} and~\ref{fig:key_token_count_full}.

\clearpage
\onecolumn
\section{Qualitative Comparison}
\label{app:qualitative}

\begin{center}
\setlength{\fboxsep}{6pt}
\fbox{%
\begin{minipage}[t][0.68\textheight][t]{\dimexpr\textwidth-2\fboxsep-2\fboxrule\relax}
\footnotesize
\raggedright
\textbf{Question:} Find the number of triples of nonnegative integers $(a,b,c)$ satisfying $a+b+c=300$ and
\[
a^2b+a^2c+b^2a+b^2c+c^2a+c^2b=6{,}000{,}000.
\]
\textbf{Ground-truth answer:} $\boxed{601}$

\vspace{0.4em}
\hrule
\vspace{0.5em}

\textbf{[GRPO]}

\texttt{<think>}
Okay, so I need to find the number of triples of nonnegative integers $(a,b,c)$ satisfy $a^2b+a^2c+b^2a+b^2c+c^2a+c^2b=6{,}000{,}000$ \\
...So that simplifies the second equation to:
$a^2(300-a)+b^2(300-b)+c^2(300-c)=6{,}000{,}000$\\
this is satisfied for any $(100,x,200-x)$, which is 201 solutions. Therefore, the answer is 201.\\
\texttt{</think>}\\
\textcolor{dneg}{Wrong answer: $\boxed{201}$}\\
\textcolor{dneg}{\textbf{Error.}} This response only counts the family with $a=100$. It misses the symmetric families where $b=100$ or $c=100$, so it undercounts the ordered triples.

\vspace{0.9em}
\hrule
\vspace{0.5em}

\textbf{[OPD]} 

\texttt{<think>}
Okay, so I need to find the number of triples of nonnegative integers (a, b, c) such that a + b + c = 300...\\
the actual \textbf{number of ordered triples} where \textbf{exactly one variable is 100} and the other two sum to 200 is: 201 total unordered $(b,c)$ pairs that result in $b+c=200$...\\
...If we consider all ordered triples where one variable is 100 and the other two are nonnegative integers summing to 200, then the number of such ordered triples is 3 * 201. Because for each of the three positions where the 100 can be placed, and for each such position, there are 201 ordered pairs (b, c).
Hence, there are \textbf{$3*201=603$ ordered triples}. \\
\texttt{</think>}\\
\textcolor{dneg}{Wrong answer: $\boxed{603}$}\\
\textcolor{dneg}{\textbf{Error.}} This response counts all three symmetric families but does not remove their overlap. The triple $(100,100,100)$ appears in all three families, so it is counted three times instead of once.

\vspace{0.9em}
\hrule
\vspace{0.5em}

\textbf{[OPD+ReNIO]} 

\texttt{<think>}
Okay, so I need to find the number of triples of nonnegative integers (a, b, c) such that a + b + c = 300 and...
Therefore, the number of solutions is the number of nonnegative integer solutions to $b+c=200$, which is 201 (from $b=0$ to $b=200$).\\
...for each of the three positions where 100 can be, and for each of the 201 possible pairs $(b,c)$, there are 3 positions for 100.\\
But wait, if we fix $a=100$, then $b+c=200$, which gives 201 solutions. Similarly, if we fix $b=100$, then $a+c=200$, giving another 201 solutions, and similarly for $c=100$, giving another 201 solutions.\\
This would lead to $3 * 201 = 603$ solutions.\\
...\\
Alternatively, since each of the three sets ($a=100$, $b=100$, $c=100$) includes the triple $(100,100,100)$ once. Therefore, the total number of distinct ordered triples is $3 * 201 - 2 = 603 - 2 = 601.$Therefore, the number of triples is 601.\\
\texttt{</think>}\\
\textcolor{dpos}{Correct answer: $\boxed{601}$}\\

\textcolor{dpos}{\textbf{Correct.}} This response preserves both necessary steps: it uses symmetry to count all ordered placements of $100$, and it applies overlap correction to avoid double counting.
\end{minipage}}
\captionof{figure}{Qualitative comparison on one mathematical counting problem. GRPO finds a valid family but misses symmetric placements; OPD handles symmetry but misses duplicate removal; OPD+ReNIO gives the correct answer by preserving both steps.}
\label{fig:qualitative-counting}
\end{center}

\clearpage

\end{document}